\documentclass{article}

\usepackage[preprint]{neurips_2026}

\usepackage[utf8]{inputenc} 
\usepackage[T1]{fontenc}    
\usepackage{hyperref}       
\usepackage{url}            
\usepackage{booktabs}       
\usepackage{amsfonts}       
\usepackage{nicefrac}       
\usepackage{microtype}      
\usepackage[table]{xcolor}
\usepackage{amsmath}    
\usepackage{amssymb}    
\usepackage{amsthm}
\usepackage{mathtools}  
\usepackage{bbm}        
\usepackage{graphicx}
\usepackage{multirow}
\usepackage{multicol}
\usepackage{colortbl}
\usepackage{caption}
\usepackage{array}
\usepackage{makecell}
\usepackage{listings}
\usepackage{paralist}

\usepackage{siunitx}
\usepackage{wrapfig}
\usepackage{subcaption}
\usepackage{algorithm}
\usepackage{longtable}
\usepackage{arydshln}
\usepackage{enumitem}
\usepackage{tabularx}
\usepackage{algpseudocode}
\usepackage{titlesec}
\usepackage{fancyhdr}

\setcitestyle{numbers,square} 
\hypersetup{
    colorlinks,
    linkcolor={blue!80!black},
    citecolor={blue!80!black},
}
\definecolor{White}{rgb}{1,1,1}
\definecolor{LightCyan}{rgb}{0.88,1,1}

\title{SEIF: Self-Evolving Reinforcement Learning \\ for Instruction Following}

\author{
 \textbf{Qingyu Ren\textsuperscript{1}},
 \textbf{Qianyu He\textsuperscript{1}},
 \textbf{Jiajie Zhu\textsuperscript{2}}, \textbf{Xingzhou Chen\textsuperscript{1}},
 \textbf{Jingwen Chang\textsuperscript{1}},\\
 \textbf{Zeye Sun\textsuperscript{3}},
 \textbf{Han Xia\textsuperscript{3}},
 \textbf{Fei Yu\textsuperscript{3}},
 \textbf{Jiaqing Liang\textsuperscript{2}},
 \textbf{Yanghua Xiao\textsuperscript{1}}\\
    \textsuperscript{\rm 1}Shanghai Key Laboratory of Data Science, \\College of Computer Science and Artificial Intelligence, Fudan University,\\
    \textsuperscript{\rm 2}School of Data Science, Fudan University,
    \textsuperscript{\rm 3}Ant Group\\
     \{qyren24,qyhe21\}@m.fudan.edu.cn, \{liangjiaqing, shawyh\}@fudan.edu.cn
}

\begin{document}

\maketitle

\begin{abstract}
Instruction following is a fundamental capability of large language models (LLMs), yet continuously improving this capability remains challenging. Existing methods typically rely either on costly external supervision from humans or strong teacher models, or on self-play training with static-difficulty instructions that cannot evolve as the model’s capabilities improve. To address these limitations, we propose \textbf{SEIF} (\textbf{S}elf-\textbf{E}volving Reinforcement Learning for \textbf{I}nstruction \textbf{F}ollowing), a self-evolving framework for enhancing the instruction-following ability of LLMs. SEIF forms a closed self-evolution loop that improves the model's instruction-following ability, where  instruction  difficulty evolution and model capability evolution reinforce each other. SEIF consists of four roles: an Instructor that generates increasingly challenging instructions, a Filter that removes conflicting or invalid instructions to ensure data quality, a Follower that learns to follow evolved instructions, and a Judger that provides reward signals for reinforcement learning. The Instructor and Follower are alternately trained and co-evolve throughout the process. Experiments across multiple model scales and architectures show that SEIF consistently improves instruction-following performance, suggesting strong generality. Further analyses reveal the sources of improvement and identify an effective training strategy for self-evolution on open-ended tasks: sufficient early-stage training to build a solid  foundation, followed by moderate late-stage training to mitigate overfitting and achieve better final performance. The code and data are publicly available at \href{https://github.com/Rainier-rq1/SEIF}{https://github.com/Rainier-rq1/SEIF}.

\end{abstract}

\section{Introduction}

Instruction following is one of the key criteria for evaluating the ability of large language models~\citep{zhang2026recommendation,pyatkin2025generalizing,shi2025hi,gou2026mixture}. A model with strong instruction-following ability can accurately understand the core requirements and constraints in user instructions, and generate responses that meet expectations~\citep{weller2025followir,yuan2025following,yuan2025easytool,zhang2025iopo}. As LLMs are widely applied in  real scenarios, how to continuously and efficiently improve the instruction-following ability of LLMs has become an important research topic~\citep{an2025ultraif,wang2026light,guo2025ifdecorator,he2024complex}.

\begin{figure}[t]
        \centering\includegraphics[width=1.0\textwidth]{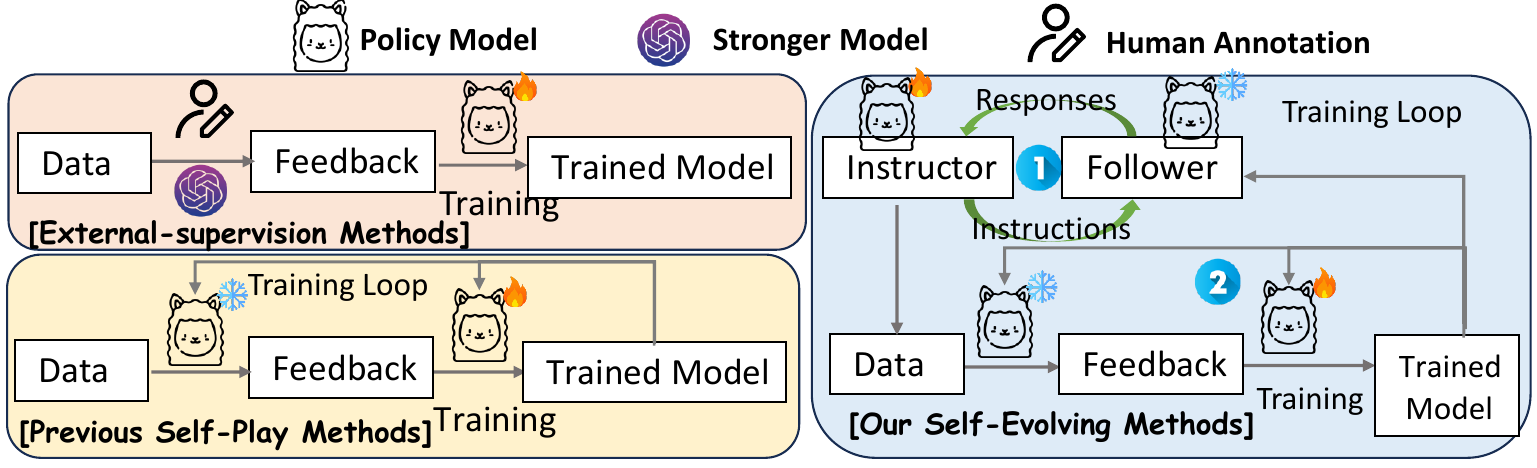}
\caption{Comparison of  training paradigms for instruction-following. Existing methods either rely on external supervision or improve through self-play with instructions of static difficulty, which cannot evolve as the model’s capabilities improve. In contrast, SEIF establishes a closed self-evolution loop during training, in which the Instructor generates  more challenging instructions, and the Follower learns to follow the new instructions, enabling co-evolution.}

        \label{fig:intro}
\end{figure}

However, improving instruction-following ability remains challenging. For improving instruction-following, there are several paradigms, as shown in Figure~\ref{fig:intro}. Specifically, some methods rely on strong teacher models or human annotations to provide feedback, which is costly and not scalable~\citep{peng2025verif,qin2025incentivizing}. Other methods improve the ability through the model's own self-play. However, the difficulty of instructions used by these methods is usually static~\citep{huang2025musc,cheng2024spar,wu2025meta}. The instructions cannot continuously evolve with the enhancement of model capability, which limits further breakthroughs in model capability. Recently, some works explore the self-evolution paradigm. Through their own iterative generation, feedback, and learning, the models continuously improve their capabilities. However, current self-evolving methods focus on verifiable tasks such as math and code~\citep{huang2025r,zhao2025absolute,chen2025self}. Self-evolution research for open-ended tasks such as instruction following has important research value.

Self-evolving instruction-following faces the following challenges. (1) \textit{Instruction Difficulty Evolution}: how to continuously generate increasingly challenging instructions as model capabilities improve? (2) \textit{Instruction Quality Assurance}: how to prevent constraint conflicts and maintain data quality as instruction complexity grows? (3) \textit{Reward Signal Acquisition}: how to obtain reliable reward signals for open-ended instruction-following tasks that do not have a ground-truth answer?

\begin{figure}[t]
        \centering\includegraphics[width=0.9\textwidth]{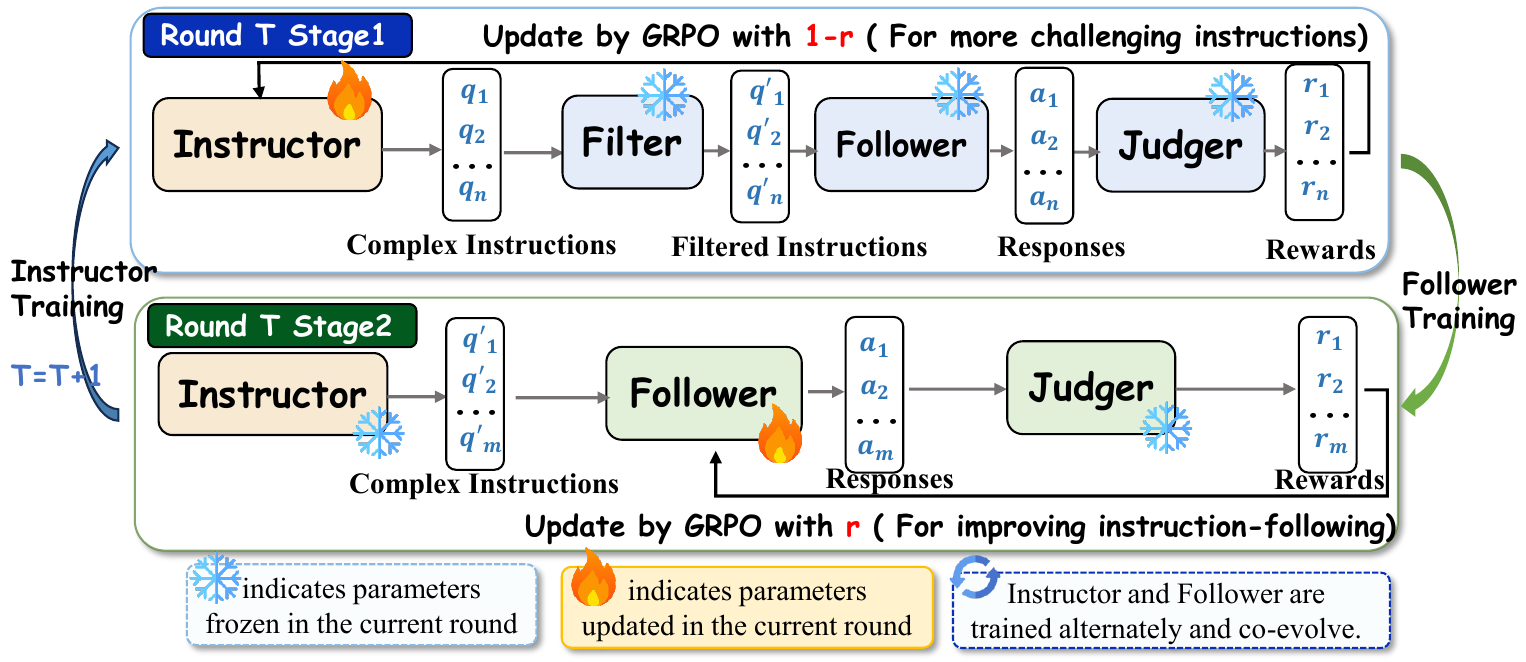}
        \caption{
Overview of the SEIF framework, which consists of two stages. The Instructor and Follower are initialized from the same base model. In Stage 1, the Instructor is updated by GRPO with reversed rewards to generate more challenging instructions for the latest Follower. In Stage 2, the Follower is updated by GRPO to follow the new instructions generated by the latest Instructor. Filter is used to filter conflicting or meaningless instructions, while Judger is used to provide rewards. The same color indicates that the models share the same initial parameters.
}
        \label{fig:method}
\end{figure}

To address these challenges, we propose \textbf{SEIF} (\textbf{S}elf-\textbf{E}volving Reinforcement Learning for \textbf{I}nstruction \textbf{F}ollowing), a self-evolving framework for improving the instruction-following ability of LLMs, as shown in Figure~\ref{fig:method}. Our framework forms a complete self-evolution loop that enhances the model's instruction-following ability by relying on the model itself. Specifically, our framework consists of four roles: Instructor, Filter, Follower, and Judger. \textit{To realize the dynamic evolution of instruction difficulty}, the Instructor generates increasingly challenging instructions tailored to the latest instruction-following model Follower. The Instructor and Follower are initialized from the  base model. \textit{To guarantee the quality of evolved instructions}, the Filter removes conflicting or invalid instructions produced during evolution. \textit{To obtain reward signals for training}, we introduce the Judger model to provide reward signals for  responses of the Follower. In our framework, the Instructor and Follower co-evolve: in each iteration, one role freezes its parameters for inference while the other is updated for training, and they alternate roles over time. Through alternating GRPO training, the Instructor pushes instructions toward the Follower's capability boundary, while the Follower improves on this evolving distribution. The Judger and Filter are instantiated from the latest Follower, enabling filtering and evaluation criteria to adapt as the model improves.

We conduct experiments across multiple parameter scales and model architectures. The results show that SEIF can improve the instruction-following ability of LLMs, demonstrating strong generalization. We further perform  analyses to investigate the sources of improvement and identify effective training strategies. The results suggest that "sufficient early-stage training followed by moderate late-stage training" is particularly effective under the self-evolution paradigm. Specifically, intensive training in the early stage helps establish a solid capability foundation, while reducing training intensity in later stages mitigates overfitting and leads to better final performance.

In summary, our contributions are three-fold: (1) We propose a self-evolving framework SEIF for improving the instruction-following ability of LLMs, offering a new perspective and practical paradigm for self-evolution training on open-ended tasks. (2) Experiment results demonstrate that our method can improve instruction-following performance across different model scales and architectures, validating its effectiveness and generalizability. (3) We conduct systematic ablations and analyses to reveal the sources of improvement and identify effective training strategies, providing useful guidance for future self-evolution research.

\section{Method}
We propose \textbf{SEIF} (\textbf{S}elf-\textbf{E}volving Reinforcement Learning for \textbf{I}nstruction \textbf{F}ollowing), a self-evolving framework for improving the instruction-following ability of large language models, as shown in Figure~\ref{fig:method}. SEIF alternates between two trainable roles: an \textit{Instructor}, which evolves seed instructions into more complex ones, and a \textit{Follower}, which learns to satisfy the generated challenging instructions. To support open-ended instruction-following training, SEIF further introduces two frozen auxiliary roles instantiated from the latest Follower: a \textit{Filter}, which removes invalid or conflicting instructions, and a \textit{Judger}, which estimates constraint satisfaction rewards.

\subsection{Overview}
\label{sec:method_overview}

Given a seed instruction $z \in \mathcal{D}_{seed}$, the Instructor $I_{\psi}$ generates an evolved instruction $x \sim I_{\psi}(\cdot \mid z)$ by adding constraints. The Filter $Q$ determines whether $x$ is conflicting or invalid, returning $Q(x) \in \{0,1\}$. An instruction is filtered if it is conflicting or invalid. Given a retained instruction $x$, the Follower $F_{\theta}$ produces a response $y \sim F_{\theta}(\cdot \mid x)$. The Judger $J$ evaluates whether $y$ satisfies the constraints in $x$ and outputs a satisfaction score $A_J(x,y) \in [0,1]$. Training proceeds in alternating stages. At iteration $t$, SEIF first uses the current Follower $F_{\theta_t}$ to instantiate frozen auxiliary roles $Q_t$ and $J_t$. The Instructor is then updated from $I_{\psi_t}$ to $I_{\psi_{t+1}}$ while keeping $F_{\theta_t}$ frozen. Next, the updated Instructor generates new instructions for Follower training, and the Follower is updated from $F_{\theta_t}$ to $F_{\theta_{t+1}}$ using rewards from the frozen Judger. In the next iteration, the auxiliary roles are refreshed from the latest Follower $F_{\theta_{t+1}}$, allowing the filtering and evaluation criteria to evolve with model capability. We provide the pseudocode of the method in Appendix~\ref{sec:pse}.

\subsection{Filter and Judger}
\label{sec:filter_judger}

SEIF introduces two frozen auxiliary roles, Filter and Judger, to ensure instruction quality and obtain rewards for open-ended instruction-following tasks. As instruction complexity increases, evolved instructions may become conflicting or invalid, which can degrade data quality and misguide training. At iteration $t$, the Filter $Q_t$, instantiated from the latest Follower $F_{\theta_t}$ and kept frozen, outputs $Q_t(x)\in\{0,1\}$ for each evolved instruction $x$, where $1$ denotes retention and $0$ denotes filtering. It gates the Instructor reward by assigning zero reward to invalid or conflicting instructions. Since open-ended instruction-following tasks lack unique ground-truth answers, the Judger $J_t$, also instantiated from $F_{\theta_t}$ and kept frozen, evaluates each instruction-response pair $(x,y)$ at the constraint level. For an instruction with $K$ constraints, it predicts a binary label $s_k\in\{0,1\}$ for each constraint and uses the satisfaction rate $A_{J_t}(x,y)=\frac{1}{K}\sum_{k=1}^{K}s_k\in[0,1]$ as the scalar reward. The Judger rewards the Instructor for generating valid instructions that the frozen Follower fails to satisfy, and rewards the Follower for producing responses that satisfy more constraints. By refreshing $Q_t$ and $J_t$ from the latest Follower, SEIF obtains adaptive filtering and reward criteria. Details are in Appendix~\ref{sec:aaaa}.

\subsection{Instructor Optimization}
\label{sec:instructor}

At iteration $t$, the Instructor $I_{\psi_t}$ generates an evolved instruction $x \sim I_{\psi_t}(\cdot \mid z)$ from a seed instruction $z \sim \mathcal{D}_{seed}$. The frozen Filter $Q_t$, instantiated from the latest Follower $F_{\theta_t}$, first checks whether $x$ is conflicting or invalid. If $Q_t(x)=0$, the instruction is filtered and receives zero reward. If $Q_t(x)=1$, the frozen Follower $F_{\theta_t}$ responds to $x$, producing $y \sim F_{\theta_t}(\cdot \mid x)$. The frozen Judger $J_t$ then estimates the satisfaction rate $A_t(x,y)=A_{J_t}(x,y)$. The Instructor is encouraged to generate instructions that are not  satisfied by the latest Follower. The Instructor reward is defined as:
\begin{equation}
    R_I(z,x) =
    \begin{cases}
        1 - A_t(x,y), & Q_t(x)=1, \\
        0, & Q_t(x)=0.
    \end{cases}
    \label{eq:instructor_reward}
\end{equation}

This reward pushes the Instructor to discover instructions near the current capability boundary of the Follower. The Filter ensures the quality of the training data, while the Judger measures how much the instruction challenges the current Follower. During this stage, only the Instructor parameters are updated, and the Follower, Filter, and Judger are kept frozen.

\subsection{Follower Optimization}
\label{sec:follower}

After obtaining the updated Instructor $I_{\psi_{t+1}}$, SEIF first uses it to generate new training instructions $x \sim I_{\psi_{t+1}}(\cdot \mid z)$, where $z \sim \mathcal{D}_{seed}$. For each  instruction $x$, the Follower samples a group of responses $\{y_i\}_{i=1}^{G} \sim F_{\theta_t}(\cdot \mid x)$. A frozen Judger $J'_t$, instantiated from the pre-update Follower $F_{\theta_t}$, evaluates each response at the constraint level. The Follower reward is:
\begin{equation}
    R_F(x,y_i) = A_{J'_t}(x,y_i).
    \label{eq:follower_reward}
\end{equation}

The Follower is optimized to maximize this reward, encouraging it to generate responses that better satisfy the constraints in the evolved instructions. Since the prompts are generated by the latest Instructor rather than sampled from a fixed dataset, the Follower is trained on an adaptive instruction distribution near its capability boundary, which drives progressive improvement.

\subsection{Reinforcement Learning Optimization}
\label{sec:grpo}

We adopt Group Relative Policy Optimization (GRPO)~\citep{guo2025deepseek} to update both trainable components in SEIF. GRPO compares several candidate outputs sampled under the same prompt and normalizes their rewards within the group, thereby avoiding the need to train an additional value function. Specifically, for a prompt $q$, a group of responses
$\{o_1,o_2,\cdots,o_G\}$ is sampled from the old policy $\pi_{\omega_{\mathrm{old}}}(\cdot|q)$. The policy $\pi_{\omega}$ is then optimized with the following clipped objective:
\begin{equation}
\small
\begin{aligned}
\mathcal{J}_{\mathrm{GRPO}}(\omega)
&=
\mathbb{E}
\left[
q \sim \mathcal{P}, \{o_i\}_{i=1}^{G} \sim \pi_{\omega_{\mathrm{old}}}(\cdot|q)
\right]
\\
&\quad
\frac{1}{G}
\sum_{i=1}^{G}
\left(
\min
\left(
\frac{\pi_{\omega}(o_i|q)}
{\pi_{\omega_{\mathrm{old}}}(o_i|q)}
A_i,
\mathrm{clip}
\left(
\frac{\pi_{\omega}(o_i|q)}
{\pi_{\omega_{\mathrm{old}}}(o_i|q)},
1-\varepsilon,
1+\varepsilon
\right)
A_i
\right)
-
\beta \mathbb{D}_{\mathrm{KL}}
\left(
\pi_{\omega}
\middle\|
\pi_{\mathrm{ref}}
\right)
\right).
\end{aligned}
\label{eq:grpo}
\end{equation}

The KL regularization term is computed with the estimator
$\mathbb{D}_{\mathrm{KL}}(\pi_{\omega}\|\pi_{\mathrm{ref}})
=
\frac{\pi_{\mathrm{ref}}(o_i|q)}{\pi_{\omega}(o_i|q)}
-
\log
\frac{\pi_{\mathrm{ref}}(o_i|q)}{\pi_{\omega}(o_i|q)}
-1$,
where $\pi_{\mathrm{ref}}$ denotes the reference policy, $\varepsilon$ is the clipping coefficient, and $\beta$ controls the strength of KL regularization. For each sampled output, the advantage is computed by standardizing its reward within the group, i.e.,
$A_i =
\frac{
r_i - \mathrm{mean}(\{r_1,r_2,\cdots,r_G\})
}{
\mathrm{std}(\{r_1,r_2,\cdots,r_G\})
}$,
where $\{r_1,r_2,\cdots,r_G\}$ are the rewards assigned to outputs in the same group. GRPO is applied to both Instructor and Follower optimization. In Instructor optimization, $q$ corresponds to the seed instruction $z$, $o_i$ corresponds to an evolved instruction $x_i$, and the reward is given by $R_I$. In Follower optimization, $q$ corresponds to the evolved instruction $x$, $o_i$ corresponds to the response $y_i$, and the reward is given by $R_F$.

\subsection{Alternating Training Procedure}
\label{sec:alternating}

At each iteration $t$, SEIF first updates the Instructor and then updates the Follower. In the Instructor stage, the current Follower $F_{\theta_t}$ is frozen and used to instantiate the Filter $Q_t$ and the Judger $J_t$. The Instructor generates evolved instructions from seed instructions, receives rewards according to Eq.~\ref{eq:instructor_reward}, and is updated by GRPO:
$
    \psi_{t+1}
    \leftarrow
    \arg\max_{\psi}
    \mathcal{J}_{\mathrm{GRPO}}^{I}(\psi).
$ In the Follower stage, the updated Instructor $I_{\psi_{t+1}}$ generates new training instructions. These instructions are used to train the Follower with rewards from the frozen Judger $J'_t$. The Follower is updated by GRPO:
$
    \theta_{t+1}
    \leftarrow
    \arg\max_{\theta}
    \mathcal{J}_{\mathrm{GRPO}}^{F}(\theta).
$ After the Follower is updated, the next iteration starts by instantiating new auxiliary roles from the latest Follower $F_{\theta_{t+1}}$. This alternating process forms a self-evolving loop: the Instructor continuously adapts the instruction distribution to the Follower's current limitations, while the Follower improves by learning from these adaptive instructions.

\section{Experiment}
\definecolor{lightblue}{RGB}{173, 216, 230}
\definecolor{lightpink}{RGB}{255, 220, 180}
\definecolor{lightgray}{RGB}{211, 211, 211}
\definecolor{lightgreen}{RGB}{144, 238, 144}
\definecolor{lightyellow}{RGB}{255, 255, 200}
\definecolor{lightred}{RGB}{255, 182, 193}
\definecolor{lightpurple}{RGB}{216, 191, 216}
\definecolor{tableblue}{RGB}{123, 166, 180}
\definecolor{darkpink}{RGB}{255, 165, 100}
\definecolor{darkgray}{RGB}{169, 169, 169}
\definecolor{darkgreen}{RGB}{94, 188, 94}
\definecolor{darkyellow}{RGB}{218, 165, 32}
\definecolor{darkred}{RGB}{205, 92, 92}
\definecolor{darkpurple}{RGB}{166, 141, 166}

\begin{table*}[t]
\centering
\caption{Performance of different models across iterative training stages on six instruction-following benchmarks. We use \textbf{bold} for the best results within each model family.}
\label{tab:main}

{\fontsize{6.8pt}{8.6pt}\selectfont
\resizebox{0.95\textwidth}{!}{%
\begin{tabular}{lccccccc}
\toprule
\multirow{2}{*}{\textbf{Models}} &
\multirow{2}{*}{\textbf{Stage}} &
\textbf{IFEval} & \textbf{CFBench} & \textbf{FollowBench}
& \textbf{WritingBench} & \textbf{AgentIF} & \textbf{Multi-IF} \\

& & \textbf{Pr.(L)} & \textbf{ISR} & \textbf{HSR}
& \textbf{Avg.} & \textbf{CSR} & \textbf{Avg.} \\

\midrule
Claude-Opus-4.7~\cite{anthropic2026claudeopus47}& --&89.1&71.0&68.9&81.0&66.1&78.2\\
GPT-4o~\cite{hurst2024gpt}
& --
& 84.8 & 65.0 & 70.4 & 75.5 & 58.5 & 71.1 \\

QwQ-32B~\citep{team2025qwq}
& --
& 83.9 & 68.0 & 62.2 & 79.1 & 58.1 & 56.2 \\

Self-Supervised-7B~\citep{ren2025instructions}
& --
& 78.9 & 52.0 & 57.5 & 58.5 & 56.7 & 64.3 \\

VERIF-8B~\citep{peng2025verif}
& --
& 87.1 & 41.0 & 56.9 & 50.8 & 56.6 & 66.2 \\

RAIF-7B~\citep{qin2025incentivizing}
& --
& 74.1 & 43.0 & 56.2 & 61.7 & 51.9 & 56.1 \\

SPAR-8B-DPO~\cite{cheng2024spar}
& --
& 82.4 & 37.0 & 56.1 & 47.0 & 53.6 & 39.6 \\

Crab-7B-DPO~\citep{qi2025constraint}
& --
& 57.7 & 25.0 & 49.4 & 45.4 & 47.2 & 31.4 \\

Conifer-7B-DPO~\citep{sun2024conifer}
& --
& 52.3 & 25.0 & 50.0 & 32.2 & 44.3 & 29.6 \\

\midrule

\multirow{4}{*}{Qwen2.5-1.5B-Instruct}
& BASE
& 43.6 & 22.0 & 34.6
& 44.8 & 42.8 & 31.7 \\

& Iter1
& 44.7 & \textbf{24.0} & 36.5
& \textbf{45.6} & 46.2 & \textbf{32.1} \\

& Iter2
& 47.1 & 23.0 & 36.0
& 45.4 & 45.6 & 31.7 \\

& \cellcolor{darkpink!25}Iter3
& \cellcolor{darkpink!25}\textbf{47.5}\textcolor{red}{(+3.9)}
& \cellcolor{darkpink!25}\textbf{24.0}\textcolor{red}{(+2.0)}
& \cellcolor{darkpink!25}\textbf{36.6}\textcolor{red}{(+2.0)}
& \cellcolor{darkpink!25}\textbf{45.6}\textcolor{red}{(+0.8)}
& \cellcolor{darkpink!25}\textbf{47.5}\textcolor{red}{(+4.7)}
& \cellcolor{darkpink!25}32.0\textcolor{red}{(+0.3)} \\

\midrule

\multirow{4}{*}{Qwen2.5-7B-Instruct}
& BASE
& 73.9 & 47.0 & 55.1
& 57.2 & 54.2 & 59.0 \\

& Iter1
& 76.5 & 50.0 & 56.6
& 63.1 & 59.3 & 61.0 \\

& Iter2
& 77.3 & 49.0 & 56.8
& 63.3 & 59.8 & 61.5 \\

& \cellcolor{tableblue!25}Iter3
& \cellcolor{tableblue!25}\textbf{78.6}\textcolor{red}{(+4.7)}
& \cellcolor{tableblue!25}\textbf{51.0}\textcolor{red}{(+4.0)}
& \cellcolor{tableblue!25}\textbf{59.0}\textcolor{red}{(+3.9)}
& \cellcolor{tableblue!25}\textbf{63.8}\textcolor{red}{(+6.6)}
& \cellcolor{tableblue!25}\textbf{60.5}\textcolor{red}{(+6.3)}
& \cellcolor{tableblue!25}\textbf{61.9}\textcolor{red}{(+2.9)} \\

\midrule

\multirow{4}{*}{Llama-3.1-8B-Instruct}
& BASE
& 73.8 & 34.0 & 53.8
& \textbf{47.5} & 53.4 & 59.5 \\

& Iter1
& 76.9 & 34.0 & 56.8
& 47.2 & 55.8 & 61.3 \\

& Iter2
& 78.0 & 35.0 & 56.9
& 47.1 & 55.2 & 61.6 \\

& \cellcolor{darkgreen!25}Iter3
& \cellcolor{darkgreen!25}\textbf{78.4}\textcolor{red}{(+4.6)}
& \cellcolor{darkgreen!25}\textbf{36.0}\textcolor{red}{(+2.0)}
& \cellcolor{darkgreen!25}\textbf{57.3}\textcolor{red}{(+3.5)}
& \cellcolor{darkgreen!25}47.4\textcolor{gray}{(-0.1)}
& \cellcolor{darkgreen!25}\textbf{57.4}\textcolor{red}{(+4.0)}
& \cellcolor{darkgreen!25}\textbf{62.4}\textcolor{red}{(+2.9)} \\

\midrule

\multirow{4}{*}{Distill-Qwen-14B}
& BASE
& 74.9 & 55.0 & 51.2
& 61.0 & 54.5 & 53.0 \\

& Iter1
& 76.4 & 58.0 & 53.2
& 60.8 & \textbf{59.7} & 57.0 \\

& Iter2
& 78.0 & 59.0 & 52.7
& 61.4 & 58.9 & \textbf{57.3} \\

& \cellcolor{darkyellow!25}Iter3
& \cellcolor{darkyellow!25}\textbf{80.0}\textcolor{red}{(+5.1)}
& \cellcolor{darkyellow!25}\textbf{60.0}\textcolor{red}{(+5.0)}
& \cellcolor{darkyellow!25}\textbf{54.0}\textcolor{red}{(+2.8)}
& \cellcolor{darkyellow!25}\textbf{62.1}\textcolor{red}{(+1.1)}
& \cellcolor{darkyellow!25}58.2\textcolor{red}{(+3.7)}
& \cellcolor{darkyellow!25}56.8\textcolor{red}{(+3.8)} \\

\midrule

\multirow{4}{*}{R1-0528-Qwen3-8B}
& BASE
& 79.7 & 66.0 & 60.4
& 76.1 & 57.4 & 48.4 \\

& Iter1
& 81.3 & 68.0 & 64.7
& 76.3 & 61.2 & 52.4 \\

& Iter2
& \textbf{81.9} & 68.0 & 64.8
& 76.1 & 62.3 & 52.6 \\

& \cellcolor{darkgray!25}Iter3
& \cellcolor{darkgray!25}\textbf{81.9}\textcolor{red}{(+2.2)}
& \cellcolor{darkgray!25}\textbf{69.0}\textcolor{red}{(+3.0)}
& \cellcolor{darkgray!25}\textbf{66.2}\textcolor{red}{(+5.8)}
& \cellcolor{darkgray!25}\textbf{76.5}\textcolor{red}{(+0.4)}
& \cellcolor{darkgray!25}\textbf{62.6}\textcolor{red}{(+5.2)}
& \cellcolor{darkgray!25}\textbf{53.2}\textcolor{red}{(+4.8)} \\

\bottomrule
\end{tabular}%
}}

\end{table*}

\subsection{Experimental Setup}
\label{sec:exp_setup}

\textbf{Models.} We evaluate \textsc{SEIF} on five models spanning from 1.5B to 14B. We compare \textsc{SEIF} against three groups of baselines: (1) frontier models; (2) specialized instruction-following models, including Self-Supervised-7B, VERIF-8B, RAIF-7B, SPAR-8B-DPO, Crab-7B-DPO, and Conifer-7B-DPO; and (3) other training baseline methods. For clarity, we name each Iter3 checkpoint as \textsc{SEIF}- followed by its model scale. For example, the Iter3 checkpoint based on Qwen2.5-7B-Instruct is denoted as \textsc{SEIF}-7B. Details are provided in Appendix~\ref{sec:baseline}.

\textbf{Data.} For training, we collected a total of 5120 seed instructions from multiple sources~\citep{kopf2023openassistant,wang2022super,wang2023self}. The constraints added to the instructions include verifiable hard constraints and semantically related soft constraints that cannot be verified using rule-based methods. Please refer to Appendix~\ref{sec: Details of Training Data} for details of the data construction process and Appendix~\ref{sec:implement} for training implementation details. For evaluation, we use six instruction-following benchmarks: IFEval~\citep{zhou2023instruction}, CFBench~\citep{zhang2025cfbench}, FollowBench~\citep{jiang2024followbench}, WritingBench~\citep{wu2025writingbench}, AgentIF~\citep{qi2025agentif}, and Multi-IF~\citep{he2024multi}. These benchmarks cover different instruction-following scenarios, including constraint satisfaction, multi-turn dialogue, writing-oriented tasks, agentic tool-use settings, and multilingual instruction following. To assess the general capabilities of the trained models, we also evaluate on several general-purpose benchmarks, including GPQA-Diamond~\citep{rein2023gpqa}, MMLU-Pro~\citep{wang2024mmlu}, BBEH~\citep{kazemi2025big}, and AIME. Details are provided in Appendix~\ref{sec:benchmark}.

\subsection{Main Results}

\textbf{Self-Evolving Enhances Instruction-Following.} Table~\ref{tab:main} demonstrates that SEIF can generally improve instruction-following capability across five model families (1.5B to 14B parameters) over three iterative training stages. On Qwen2.5-7B-Instruct, SEIF achieves $+4.7$ on IFEval, $+4.0$ on CFBench, and $+6.6$ on WritingBench; Distill-Qwen-14B reaches $80.0$ on IFEval ($+5.1$). Our self-evolving approach outperforms other instruction-following optimization baselines, confirming that self-evolution is effective for enhancing instruction-following capability.

\textbf{Small Models Can Also Benefit from Self-Evolution.} As shown in Table~\ref{tab:main}, even the smallest model,
Qwen2.5-1.5B-Instruct, achieves consistent gains from the proposed
self-evolution process. The largest improvements are observed on AgentIF
($+4.7$) and IFEval ($+3.9$), suggesting that self-evolution effectively
enhances small models' instruction-following ability. Although its absolute
scores remain lower than those of larger models, the steady gains across
nearly all benchmarks show that small-scale models are not solely constrained
by parameter size, but can still develop stronger instruction-following
behaviors through iterative self-training.

\textbf{Instruction Difficulty Evolution is Critical.} Table~\ref{tab:baseline} compares SEIF against other baselines on SEIF-7B. Static instruction self-play methods (Self-Correct, Humpback, SELF, Self-Rewarding, I-SHEEP, Meta-Rewarding) show only marginal improvements over the base model; the strongest baseline (Meta-Rewarding) achieves 76.6 on IFEval, only $+2.7$ over BASE. In contrast, SEIF with dynamic instruction difficulty evolution achieves the best performance: $78.6$ on IFEval, $51.0$ on CFBench, and $59.0$ on FollowBench, representing a $+2.0$ improvement over Meta-Rewarding on IFEval. The variant without Instructor Evolving shows a significant drop to $75.9$ on IFEval, confirming that self-play alone is insufficient; the \textit{adaptiveness} of the training data distribution is crucial.

\textbf{General Capability Preservation.} Table~\ref{tab:general} verifies that SEIF generally maintains model general abilities. SEIF achieves competitive or improved performance on these benchmarks: Distill-Qwen-14B reaches the best average of $60.3$, and Llama-3.1-8B improves from $24.0$ to $24.4$. This confirms that our self-evolution method does not compromise the model's general capabilities generally.

\begin{table*}[t]

  \caption{Comparison with other baseline methods on SEIF-7B.}
  \centering
  \label{tab:baseline}
  \setlength{\tabcolsep}{5pt}
  \renewcommand{\arraystretch}{1.12}
  \resizebox{0.9\linewidth}{!}{%
    \begin{tabular}{@{}llccc@{}}
      \toprule
      \textbf{Model} 
      & \textbf{Mechanism}
      & \textbf{IFEval Pr.(L)} 
      & \textbf{CFBench ISR} 
      & \textbf{FollowBench HSR} \\
      \midrule
      BASE                         & Base model inference                          & 73.9 & 47.0 & 55.1 \\
      SFT                          & Supervised fine-tuning       & 74.2 & 46.0 & 55.7 \\
       ProxyReward~\cite{guo2025general}                  & Targeted proxy reward        & 76.1 & 49.0 & 57.0 \\
      \hdashline
      Self-Correct~\cite{ferraz2024llm}               & Decompose-critique-refine    & 73.4 & 46.0 & 54.8 \\
      Humpback~\cite{li2023self}                     & Instruction backtranslation  & 73.7 & 47.0 & 55.3 \\
      SELF~\cite{lu2023self}                  & Language-feedback evolution  & 76.2 & 47.0 & 56.1 \\
      Self-Rewarding~\citep{yuan2024self}                & LLM-as-a-judge reward        & 76.4 & 48.0 & 56.7 \\
      I-SHEEP~\cite{liang2024sheep}               & Generate-assess-filter       & 75.8 & 49.0 & 56.4 \\
      Meta-Rewarding~\citep{wu2025meta}               & LLM-as-a-meta-judge          & 76.6 & 49.0 & 57.6 \\
      \hdashline
      Ours  (w/o Instructor Evolving) & Static Instruction Difficulty           & 75.9 & 48.0 & 56.1 \\
      \rowcolor{blue!10}
      \textbf{Ours-SEIF}
      & \textbf{Instructor--Follower co-evolution} & \textbf{78.6} & \textbf{51.0} & \textbf{59.0} \\
      \bottomrule
    \end{tabular}
  }%
\end{table*}

\begin{table*}[t]
\centering
\caption{Performance on general benchmarks. We use \textbf{bold} for the best results and \underline{underlined} for the second-best results. We evaluate AIME using the Avg@30 method.
}
\label{tab:general}
\renewcommand{\arraystretch}{1.12}
\resizebox{0.9\textwidth}{!}{
\begin{tabular}{llccccc>{\columncolor{blue!10}}c}
\toprule
\textbf{Model} & \textbf{Method} & \textbf{GPQA-Diamond} & \textbf{MMLU-Pro} & \textbf{BBEH} & \textbf{AIME24} & \textbf{AIME25} & \textbf{Avg.} \\
\midrule
\multirow{3}{*}{Qwen2.5-7B-Instruct}
& BASE & 32.3 & 46.9 & 40.5 & 10.9 & 7.7 & 27.7 \\
& w/o Instructor Evolving & 35.4 & 48.3 & 42.5 & 10.3 & 7.3 & \textbf{28.8} \\
& Ours-SEIF  & 32.8 & 48.3 & 40.5 & 11.3 & 7.3 & \underline{28.0} \\
\midrule
\multirow{3}{*}{Llama-3.1-8B-Instruct}
& BASE & 30.3 & 43.0 & 42.5 & 3.8 & 0.6 & \underline{24.0} \\
& w/o Instructor Evolving & 25.8 & 43.7 & 43.5 & 4.4 & 0.4 & 23.6 \\
& Ours-SEIF & 28.8 & 43.4 & 44.5 & 5.0 & 0.4 & \textbf{24.4} \\
\midrule
\multirow{3}{*}{Distill-Qwen-14B}
& BASE & 56.1 & 71.1 & 52.5 & 67.8 & 49.4 & \underline{59.4} \\
& w/o Instructor Evolving & 53.5 & 71.2 & 51.5 & 68.7 & 49.1 & 58.8 \\
& Ours-SEIF & 60.1 & 71.4 & 52.0 & 68.9 & 49.0 & \textbf{60.3} \\
\bottomrule
\end{tabular}
}
\end{table*}

\section{Analysis}
\subsection{Ablation Study}
\label{sec:ablation}

Table~\ref{tab:ablation} presents ablation results on SEIF-7B by removing three key components. First, \textbf{w/o Filter} leads to clear drops on IFEval ($-3.2$) and CFBench ($-6.0$), showing that filtering invalid or conflicting instructions is crucial for maintaining data quality. Without the Filter, the Instructor may generate conflicting or invalid instructions, producing noisy training signals. The large drop on CFBench further highlights the importance of ensuring instruction feasibility and consistency. Second, \textbf{w/o Shared Parameters}, which keeps the Judger and Filter fixed instead of refreshing them from the latest Follower, decreases performance on these benchmarks. This suggests that adaptive filtering and evaluation criteria are important for keeping reward signals aligned with the evolving model capability. Third, \textbf{w/o Const.-Level Reward}, which uses instruction-level binary rewards, assigning 1 only if all constraints are satisfied. Its performance drop highlights the need of fine-grained rewards.

\begin{figure}[t]
    \centering
    \scriptsize

    \begin{minipage}[t]{0.47\linewidth}
    \centering
    \setlength{\tabcolsep}{1.2mm}
    \renewcommand{\arraystretch}{1.3}
    \captionof{table}{Ablation study results on SEIF-7B.}
    \label{tab:ablation}
    \resizebox{\linewidth}{!}{
    \begin{tabular}{lccc}
        \toprule[1pt]
        \textbf{Method} & \textbf{IFEval} & \textbf{CFBench} & \textbf{FollowBench} \\
        \midrule
        BASE & 73.9 & 47.0 & 55.1 \\
        \rowcolor{blue!10}
        \textbf{Ours-SEIF} & \textbf{78.6} & \textbf{51.0} & \textbf{59.0} \\
        \hdashline
        w/o Filter 
            & \textcolor{red}{$-3.2$}
            & \textcolor{red}{$-6.0$}
            & \textcolor{red}{$-4.1$} \\
        w/o Shared Parameters 
            & \textcolor{red}{$-1.8$}
            & \textcolor{red}{$-1.0$}
            & \textcolor{red}{$-1.9$} \\
        w/o Const.-Level Reward 
            & \textcolor{red}{$-2.6$}
            & \textcolor{red}{$-2.0$}
            & \textcolor{red}{$-1.8$} \\
        \bottomrule[1pt]
    \end{tabular}
    }
\end{minipage}
    \hfill
    \begin{minipage}[t]{0.45\linewidth}
        \centering
        \setlength{\tabcolsep}{0.7mm}
        \renewcommand{\arraystretch}{1.0}
        \captionof{table}{IFEval performance across turns.}
        \label{tab:turn}
        \resizebox{\linewidth}{!}{
        \begin{tabular}{lccc}
            \toprule[1pt]
            \textbf{Turn} & \textbf{Qwen2.5-7B} & \textbf{Llama-3.1-8B} & \textbf{Distill-Qwen-14B} \\
            \midrule
            BASE  & 73.9 & 73.8 & 74.9 \\
            Turn1 & 76.5 & 76.9 & 76.4 \\
            Turn2 & 77.3 & 78.0 & 78.0 \\
            \rowcolor{blue!10}
            Turn3 & 78.6 & 78.4 & 80.0 \\
            \hdashline
            Turn4 & 78.8\textcolor{red}{(+0.2)} & 78.5\textcolor{red}{(+0.1)} & 79.8\textcolor{gray}{(-0.2)} \\
            Turn5 & 78.6\textcolor{red}{(+0.0)} & 78.7\textcolor{red}{(+0.3)} & 79.9\textcolor{gray}{(-0.1)} \\
            \bottomrule[1pt]
        \end{tabular}
        }
    \end{minipage}
\end{figure}

\textbf{How many training iterations are needed?} Table~\ref{tab:turn} presents an ablation study on the number of iteration training turns. The performance of all models improves  during the first few turns and becomes  saturated after Turn3. Further training brings only marginal gains for some models and even leads to performance degradation. Therefore, we adopt three iteration turns in SEIF.

\subsection{Training Analysis}
\label{sec:training_strategy}

\begin{figure*}[t]
        \centering\includegraphics[width=\textwidth]{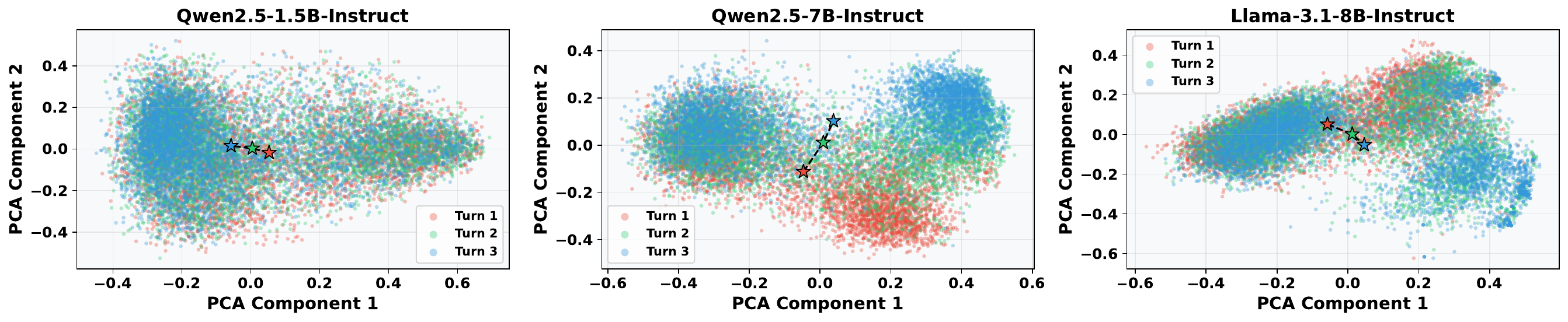}
        \caption{PCA visualization of training data representations across three iteration turns. 
        The star marker denotes the centroid of training data representations for each turn.}
        \label{fig:pca}
\end{figure*}

\begin{figure*}[t]
        \centering\includegraphics[width=0.9\textwidth]{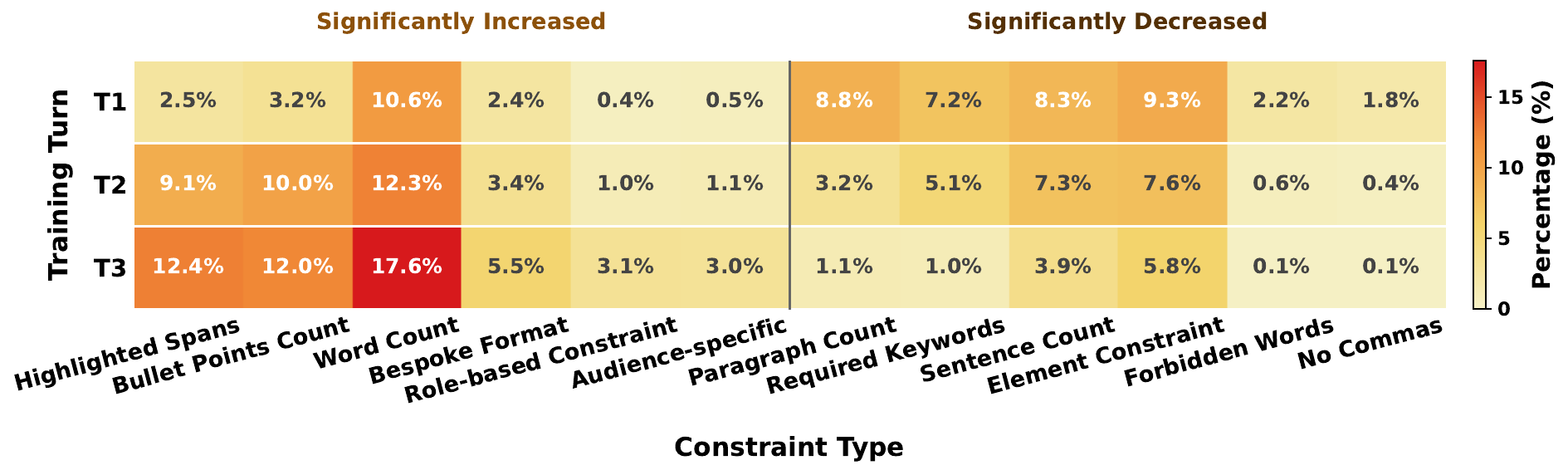}
        \caption{Top-6 constraint types with the largest proportional increases and decreases across the iteration turns of SEIF-7B. As training progresses, the constraint distribution gradually shifts from simple requirements toward more complex, structured, and semantically relevant constraint types.}
        \label{fig:heatmap}
\end{figure*}

\textbf{Why does instruction-following performance improve?} Figure~\ref{fig:pca} and Figure~\ref{fig:heatmap} provide qualitative evidence revealing the underlying mechanism of SEIF's effectiveness. Figure~\ref{fig:pca} shows PCA visualizations of training instruction representations across three iteration turns using all-MiniLM-L6-v2. We observe that the representations of training data change across iterative turns, with noticeable shifts in the distribution and centroid positions. This suggests that SEIF does not simply reuse a fixed instruction distribution, but continuously updates the training data distribution during self-evolution. The partially overlapping yet distinguishable clusters indicate that later iterations preserve semantic continuity while introducing new variations in instruction representations. Figure~\ref{fig:heatmap} further demonstrates this evolution at the constraint-type level. Simple paragraph-level or sentence-level requirements show declining proportions, while complex, structured, and semantically deep constraint types become more prevalent, including formatting constraints, role-based  constraints, and audience-specific constraints. This constraint evolution is driven by the Instructor's learning objective for generating valid instructions that the current Follower cannot satisfy. As the Follower improves, the Instructor's reward  pushes it toward increasingly challenging constraints. We provide examples showing how our instructions evolve across training iterations in Appendix~\ref{sec:iee}.

\textbf{Self-Evolving Training Strategy.}
Table~\ref{tab:train_strategies} compares six epoch allocation strategies on SEIF-7B, where $E_t=i$ denotes that training turn $t$ is trained for $i$ epochs. The results show that epoch allocation across self-evolving turns is critical: uniform schedules such as $E_1=1,E_2=1,E_3=1$ and $E_1=2,E_2=2,E_3=2$ are generally inferior to asymmetric ones, indicating that different turns contribute unequally to model improvement. A key observation is that excessive late-stage training can hurt generalization. Compared with the early-intensive schedule $E_1=3,E_2=1,E_3=1$, the late-intensive schedule $E_1=1,E_2=1,E_3=3$ shows a performance drop, suggesting that over-optimizing on later evolved data may introduce distributional bias or weaken complex instruction-following abilities. In contrast, emphasizing the first turn allows the model to sufficiently absorb foundational self-evolved data, while later turns serve as moderate refinement. Overall, $E_1=3,E_2=1,E_3=1$ achieves the best performance, demonstrating that an effective self-evolving schedule should follow the principle of ``sufficient early-stage training, followed by moderate late-stage training'' rather than using uniform or late-intensive epoch allocation.

\begin{table}[t]
\centering
\caption{Performance comparison of SEIF-7B under different training strategies. $E_t=i$ indicates that Iteration Turn $t$ was trained for $i$ epochs.}
\centering
\label{tab:train_strategies}
\setlength{\tabcolsep}{5pt}
\renewcommand{\arraystretch}{1.0}
\resizebox{0.88\linewidth}{!}{
\begin{tabular}{lcccccc}
\toprule
\textbf{Setting} & \textbf{IFEval} & \textbf{CFBench} & \textbf{FollowBench} & \textbf{Multi-IF} & \textbf{AgentIF} & \textbf{WritingBench} \\
\midrule
BASE & 73.9 & 47.0 & 55.1 & 59.0 & 54.2 & 57.2 \\
\hdashline
$E_1=1, E_2=1, E_3=1$ & 77.4 & 48.0 & 55.5 & 60.7 & 58.2 & 62.3 \\
$E_1=2, E_2=2, E_3=2$ & 77.1 & 50.0 & 57.7 & 61.3 & 58.5 & 63.0 \\
$E_1=3, E_2=3, E_3=3$ & 75.8 & 49.0 & 55.9 & 59.6 & 57.1 & 62.6 \\
\hdashline
$E_1=1, E_2=1, E_3=3$ & \textbf{78.6} & 50.0 & 56.7 & 60.8 & 60.4 & 63.0 \\
$E_1=1, E_2=3, E_3=1$ & 76.7 & 50.0 & 56.5 & 61.2 & 59.9 & 62.9 \\
\rowcolor{blue!10}
$E_1=3, E_2=1, E_3=1$ & \textbf{78.6} & \textbf{51.0} & \textbf{59.0} & \textbf{61.9} & \textbf{60.5} & \textbf{63.8} \\
\bottomrule
\end{tabular}
}
\end{table}

\begin{figure*}[t]
        \centering\includegraphics[width=0.9\textwidth]{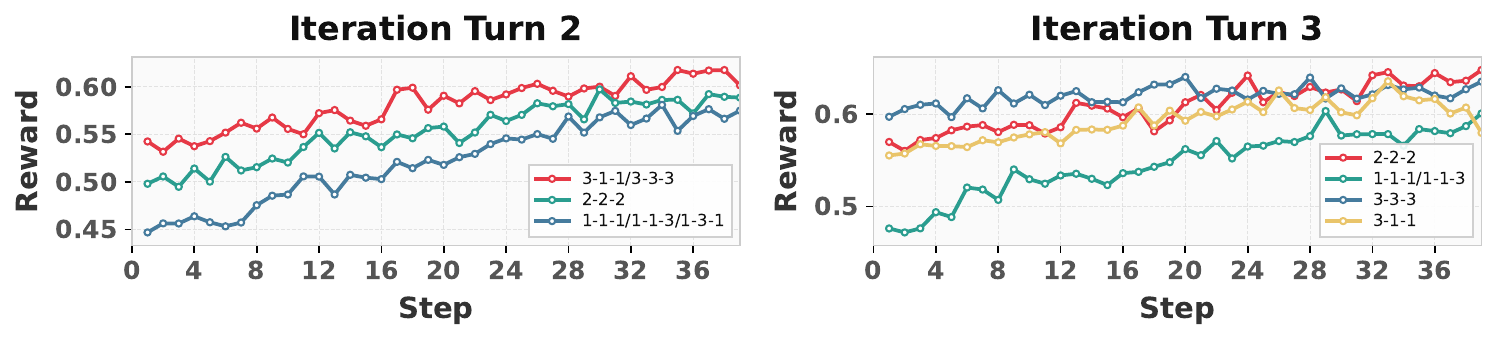}
        \caption{Reward dynamics of different training strategies across different iteration turns on SEIF-7B. X-Y-Z denotes the number of training epochs for Turn 1, Turn 2, and Turn 3, respectively. }
        \label{fig:reward}
\end{figure*}

\textbf{Why is sufficient early-stage training, followed by moderate late-stage training, a good self-evolution training strategy?} Figure~\ref{fig:reward} analyzes the reward dynamics of different training schedules. Strategies with more early-stage training, such as 3-1-1, generally achieve higher rewards and exhibit steady reward growth across Turn 2 and Turn 3, indicating that sufficient early training helps the model build a stronger foundation for subsequent self-evolution. In contrast, strategies that emphasize late-stage training, such as 1-1-3, tend to obtain lower rewards and show less favorable reward dynamics, especially in Turn 2. This suggests that excessive reliance on later-stage training may bias the model toward recently evolved instruction patterns, limiting the benefits accumulated from earlier iterations. Balanced strategies such as 2-2-2 achieve competitive rewards, but they do not consistently outperform front-loaded schedules. These results indicate that the effectiveness of self-evolution depends not only on the total training budget, but also on how the budget is allocated across iterations. A stronger early-stage foundation followed by moderate late-stage refinement improves reward stability and supports more effective self-evolution.

\textbf{Judger and Filter Reliability.}
We evaluate our self-instantiated Filter/Judger through human agreement analysis, source ablation, and final-output human evaluation. On a 400-example test set from VerInstruct~\citep{peng2025verif}, the Filter achieves stable agreement with human annotations across three iterations, with Accuracy around 0.79--0.80 and F1 around 0.78--0.80 (Table~\ref{tab:filter_results}), indicating reliable detection of conflicting instructions. The Judger obtains Accuracy around 0.73--0.74 and F1 around 0.70--0.72 (Table~\ref{tab:judger_results}), which is reasonable given the ambiguity of constraint-satisfaction judgment. Both modules show only small fluctuations across turns, suggesting no obvious reliability degradation from self-instantiation. We further show that refreshing the Filter and Judger from the latest Follower yields the best performance (Table~\ref{tab:filter_judger_source_ablation}), and blind pairwise human evaluation confirms that SEIF outputs are preferred over BASE and training baselines (Table~\ref{tab:human_pairwise_eval}).
These results support the feasibility of using self-instantiated models for automatic supervision. See Appendix~\ref{sec:H} for details.

\section{Related Work}
\subsection{Reinforcement Learning for Improving Instruction Following}

Reinforcement learning has been widely adopted to improve the instruction-following ability of LLMs. Existing methods typically derive reward signals from human annotations or stronger teacher models~\citep{peng2025verif,sun2024conifer,qin2025incentivizing}. While effective, such external supervision is costly to obtain and difficult to scale. Other work improves instruction following through self-play, where the model learns from self-generated feedback~\citep{wu2025meta,huang2025musc,cheng2024spar,yuan2024self}. However, the training data produced by these methods is often static: as the model becomes stronger, the instructions do not continuously adapt to its evolving capability, which limits further improvement. In contrast, our framework dynamically generates increasingly challenging instructions tailored to the latest Follower model.

\subsection{Self-Evolving Training for Large Language Models}

Self-evolving training aims to enable LLMs to improve through iterative data generation, feedback, and learning, while reducing dependence on external supervision. Recent studies have shown promising results in verifiable domains such as mathematical reasoning and code generation~\citep{huang2025r,zhao2025absolute,chen2025self}, where rewards can be obtained from automatic verifiers. However, open-ended instruction following is more challenging because there is no unique ground-truth answer~\citep{jiang2024followbench,zhang2025cfbench,wen2026if,pan2026rubriceval}. To address these challenges, SEIF introduces four roles. This enables instruction generation, quality assurance, and reward evaluation to co-evolve with the instruction-following capability.

\section{Conclusion}

We propose \textbf{SEIF}, a self-evolving framework for improving the instruction-following ability of LLMs. SEIF constructs a closed loop with four roles: an Instructor that evolves challenging instructions, a Filter that ensures instruction quality, a Follower that learns to follow the evolved instructions, and a Judger that provides constraint satisfaction rewards. Through alternating GRPO optimization, SEIF adapts the instruction distribution to the Follower's capability boundary and improves instruction-following ability without relying on external teacher models or human annotations during self-evolution. Experiments across different model scales and architectures demonstrate the effectiveness and generalizability of SEIF. We also find that sufficient early-stage training followed by moderate late-stage training leads to better final performance. These results suggest that self-evolving training is a promising direction for open-ended tasks.

\bibliography{ref}
\bibliographystyle{plainnat}

\newpage
\appendix

\section*{Appendix}

\section{Training Data Construction}
\label{sec: Details of Training Data}

\subsection{Pipeline}
For the \textbf{Instructor}, the input is a seed instruction, and the model output is a more complex instruction with additional constraints. The Filter model removes conflicting or invalid instructions, and such instructions directly receive a reward of 0. If an instruction passes the Filter check, the latest Follower model generates a response to this complex instruction. The Judger then computes the constraint satisfaction rate of the response, and $1$ minus the constraint satisfaction rate is used as the reward. This encourages the Instructor to generate more challenging instructions. For the \textbf{Follower}, the latest trained Instructor is used to add constraints to seed instructions, producing complex instructions as inputs. The Follower then generates responses to these complex instructions. The Judger model computes the constraint satisfaction rate of each response, which is used as the reward to encourage the Follower to better follow complex instructions. The constraints include both soft constraints and hard constraints. For each seed instruction, we construct either a soft-constrained or hard-constrained evolved instruction. The prompt template used by the model to add constraints to seed instructions is shown in Table~\ref{tab:prompt-soft}.

\subsection{Soft Constraint Types}

We categorize soft constraints into the following 25 types. Instructor can select five types of soft constraints from the constraint taxonomy and add them to the instruction. 

\paragraph{1. \textit{Content constraints}:}
These constraints require the inclusion of specific terms, symbols, or expressions, often with precise placement requirements. For example, an output may be required to include the word ``beautiful.''

\paragraph{2. \textit{Element constraints}:}
These constraints require the inclusion of specific entities, objects, events, or scenarios, such as highlighting the Great Wall.

\paragraph{3. \textit{Semantic constraints}:}
These constraints specify the intended theme, topic, tone, stance, or meaning of the response, such as writing a poem about London.

\paragraph{4. \textit{Word count constraints}:}
These constraints limit the number of words, such as requiring a 50-word poem.

\paragraph{5. \textit{Sentence count constraints}:}
These constraints limit the number of sentences, such as requiring three sentences.

\paragraph{6. \textit{Paragraph count constraints}:}
These constraints limit the number of paragraphs, such as requiring the response to be divided into three sections.

\paragraph{7. \textit{Document count constraints}:}
These constraints limit the number of documents or items, such as requiring a list of three articles.

\paragraph{8. \textit{Tone and emotion constraints}:}
These constraints require the response to adopt a particular emotional tone or attitude, such as writing a letter in an angry and sarcastic tone.

\paragraph{9. \textit{Form and style constraints}:}
These constraints specify a particular stylistic form, genre, or mode of presentation, such as writing in an encyclopedic style.

\paragraph{10. \textit{Audience-specific constraints}:}
These constraints tailor the response to a specific audience group, such as writing a poem for a 6-year-old child.

\paragraph{11. \textit{Authorial style constraints}:}
These constraints require the response to emulate the style of a particular author, such as Shakespeare.

\paragraph{12. \textit{Fundamental format constraints}:}
These constraints require standard formats such as JSON or HTML.

\paragraph{13. \textit{Bespoke format constraints}:}
These constraints impose custom formatting protocols, such as bolding the main idea and using an unordered list.

\paragraph{14. \textit{Specialized format constraints}:}
These constraints are tailored to specific applications or domains, such as converting content into an electronic medical record format.

\paragraph{15. \textit{Pragmatic constraints}:}
These constraints adapt the output to contextual, linguistic, or policy-related requirements, such as producing English or classical Chinese.

\paragraph{16. \textit{Syntactic constraints}:}
These constraints require specific phrase, clause, or sentence structures, such as using imperatives with nouns and verb phrases.

\paragraph{17. \textit{Morphological constraints}:}
These constraints regulate affixes, roots, capitalization, or word formation, such as requiring all content to be written in lowercase English.

\paragraph{18. \textit{Phonological constraints}:}
These constraints focus on sound patterns, tone, pronunciation, or intonation, such as producing single-syllable tongue twisters.

\paragraph{19. \textit{Role-based constraints}:}
These constraints require the response to adopt a specific role identity, such as answering as Confucius.

\paragraph{20. \textit{Task-specific constraints}:}
These constraints address defined situational tasks, such as reporting progress while working from home.

\paragraph{21. \textit{Complex context constraints}:}
These constraints involve multi-faceted, nested, or context-dependent reasoning, such as reasoning about a situation with ten items on the left.

\paragraph{22. \textit{Example constraints}:}
These constraints require the response to conform to patterns demonstrated by provided input--output examples.

\paragraph{23. \textit{Inverse constraints}:}
These constraints narrow the response space through exclusions, such as avoiding political topics.

\paragraph{24. \textit{Contradictory constraints}:}
These constraints combine requirements that are difficult or impossible to satisfy simultaneously, such as requesting a five-character quotation of 1000 words.

\paragraph{25. \textit{Rule constraints}:}
These constraints require the response to follow symbolic, logical, or operational rules, such as applying nonstandard arithmetic rules where each answer adds one to the usual result.

\subsection{Hard Constraint Types}

We organize the constraints into five high-level categories. Each category contains several constraint subtypes that specify particular forms of restriction on the model response, with a total of 24 subtypes. During constraint generation, Instructor first selects three different high-level categories from the taxonomy. For each selected category, it then chooses exactly one constraint subtype and generates one complete constraint based on that subtype. 

\paragraph{Category A: Lexical constraints.}
Lexical constraints regulate the occurrence of specific words, letters, or lexical forms in the response. This category includes the following subtypes.

\textit{Required keywords}: requires the response to include one or more specified keywords.

\textit{Keyword frequency}: requires a particular keyword to appear a specified number of times.

\textit{Forbidden words}: prohibits the response from containing certain specified words.

\textit{Letter frequency}: requires a specified letter to appear a certain number of times.

\textit{ALL-CAPS word frequency}: requires words written entirely in capital letters to appear a specified number of times.

\paragraph{Category B: Structural layout constraints.}
Structural layout constraints control the overall organization and length of the response. This category includes the following subtypes.

\textit{Sentence count}: requires the response to contain a specified number of sentences.

\textit{Paragraph count}: requires the response to contain a specified number of paragraphs.

\textit{Bullet points count}: requires the response to contain a specified number of bullet points.

\textit{Sectioned structure}: requires the response to be organized into multiple sections.

\textit{Nth paragraph first word}: requires the first word of a specified paragraph to be a given word.

\textit{Word count}: requires the response to satisfy a specified word-count condition.

\paragraph{Category C: Formatting constraints.}
Formatting constraints specify how the response should be rendered or wrapped. This category includes the following subtypes.

\textit{Highlighted spans}: requires certain parts of the response to be highlighted using a specified markup format.

\textit{Title wrapper}: requires the response to include a title wrapped in a specified format.

\textit{Quotation wrapper}: requires the entire response or a specified part of it to be wrapped in quotation marks.

\textit{JSON output}: requires the entire response to be valid JSON.

\paragraph{Category D: Language  constraints.}
Language constraints regulate the language, capitalization, or punctuation of the response. This category includes the following subtypes.

\textit{Response language}: requires the response to be written in a specified language and no other language.

\textit{English ALL CAPS}: requires the response to be written in English using only uppercase letters.

\textit{English all lowercase}: requires the response to be written in English using only lowercase letters.

\textit{No commas}: prohibits the use of commas in the entire response.

\paragraph{Category E: Special pattern constraints.}
Special pattern constraints impose global response patterns or require the response to contain special components. This category includes the following subtypes.

\textit{Repeat-then-answer}: requires the response to first repeat the original request and then provide the answer.

\textit{Exact ending phrase}: requires the response to end with a specified phrase, with no additional content after it.

\textit{Two distinct responses}: requires the response to contain two different answers, usually separated by a specified delimiter.

\textit{Postscript}: requires the response to include a postscript at the end, introduced by a specified marker.

\textit{Placeholder count}: requires the response to contain a specified number of placeholders in a given format.

\begin{table*}[t]
\centering
\caption{Prompt template for adding  constraints.}
\label{tab:prompt-soft}
\renewcommand{\arraystretch}{1.2}
\small
\setlength{\tabcolsep}{6pt}
\begin{tabular}{p{0.97\textwidth}}
\hline

\textbf{[Task Description]} \\

1. I currently have a seed question, but the seed question is relatively simple. To make the instruction more complex, I want you to identify and return  atomic constraints that can be added to the seed question. \\
2. I will provide \textbf{[Seed Question]} and \textbf{[Constraint References]}. You may use these references to propose  constraints that increase the difficulty of the seed question. \\
3. \textbf{[Constraint References]} are only suggestions. You may choose one or more constraints from the list, or propose new constraints if needed. \\
4. Do not modify, rewrite, or answer the seed question. Your task is only to generate additional constraints that can be added to it. \\
5. Each added constraint should be atomic, specific, and verifiable. Avoid vague, redundant, or overlapping constraints. \\
6. Return the added constraints in the following JSON format: \\
\quad \texttt{json} \\
\quad \texttt{\{} \\
\quad\quad \texttt{"c1": "<first constraint>",} \\
\quad\quad \texttt{"c2": "<second constraint>",} \\
\quad\quad $\ldots$\\
\quad \texttt{\}} \\
7. Do not return anything else. No explanation, no reformulated question, no analysis---only the JSON structure. \\

\textbf{[Constraint Type References]} \\
1. Lexical content constraint: \{Definition\} \{Example\} \\
$\ldots$ \\
25. Rule Constraint: \{Definition\} \{Example\} \\

\textbf{[Seed Question]} \\
\{raw\_question\} \\

\hline
\end{tabular}
\end{table*}

\section{Training Details}
\label{sec:implement}
We implement GRPO training of Instructor and Follower using the EasyR1 framework. All hyperparameters used in our experiments are summarized in Table~\ref{tab:aaaa}.
\begin{table}[htbp]
\centering
\small
\begin{minipage}{0.65\linewidth}
\caption{GRPO training hyperparameters.}
\label{tab:aaaa}
\resizebox{\linewidth}{!}{%
\begin{tabular}{l c}
\toprule
\textbf{Hyperparameter} & \textbf{Value} \\
\midrule
Training Framework & EasyR1~\cite{zheng2025easyr1} \\
\midrule
Global Batch Size & 96 \\
Micro Batch Size Per Device For Update & 4 \\
Micro Batch Size Per Device For Experience & 8 \\
Use KL Loss & True\\
Rollout Batch Size & 384 \\
Rollout n & 5 \\
Max Prompt Length & 2048 \\
Max Response Length & 8192 \\
\midrule
Training Steps & \\
\quad Instructor T1/T2/T3 & 13/13/13 \\
\quad Follower T1/T2/T3 & 39/13/13 \\
\midrule
Actor Learning Rate & $1 \times 10^{-6}$ \\
KL Coefficient & $1 \times 10^{-2}$ \\
\midrule
Number of GPUs & 
\begin{tabular}[c]{@{}c@{}}
8 H200 for Training \\
4 H200 for vLLM Service
\end{tabular}
\\
\bottomrule

\end{tabular}%
}
\end{minipage}
\end{table}

\section{Pseudocode}
\label{sec:pse}
We present the pseudocode of SEIF in Algorithm~\ref{alg:seif}.

\section{Filter and Judger Prompt}
\label{sec:aaaa}
We present the Filter prompt in Table~\ref{tab:conflict-checker-prompt-main}. We present the Judger prompt in Table~\ref{tab:constraint-checking-query-template}.

\begin{table}[htbp]
\centering
\caption{Prompt for Filter filtering.}
\label{tab:conflict-checker-prompt-main}

\begin{minipage}{1.0\textwidth}

\hrule

\textbf{Role.}\\
You are an \textbf{instruction-constraint consistency checker}. Determine whether a given set of instructions or constraints contains an \textbf{internal logical conflict}, i.e., whether all requirements can be satisfied simultaneously without adding extra information and while strictly following the constraints.
\\

\textbf{Core Principles.}\\
Judge a case as conflicting only when there is a provable logical contradiction or mutually exclusive constraint.
Difficulty, tight word limits, high information density, or poor/incomplete writing are feasibility challenges, not conflicts.
If at least one possible output can satisfy all constraints, output \texttt{[[1]]}.
Output \texttt{[[0]]} only when at least one pair of constraints is formally mutually exclusive.
\\

\textbf{Conflict Criteria.}\\
A conflict should be judged if any of the following holds:
\begin{itemize}[leftmargin=*, itemsep=0pt, topsep=0pt]
    \item \textbf{Mutually exclusive attribute conflict}: the same output attribute must take incompatible values, e.g., all lowercase while requiring an uppercase word.
    \item \textbf{Structural or format conflict}: required structures cannot coexist, e.g., one sentence and three paragraphs.
    \item \textbf{Language conflict}: mutually exclusive language or character rules, e.g., entirely Chinese and lowercase English.
    \item \textbf{Exact phrase matching conflict}: a fixed phrase is required while another global rule changes its capitalization, symbols, or format.
    \item \textbf{Impossible quantity or source constraint}: required items cannot be extracted from the input while external knowledge or new entities are prohibited.
\end{itemize}

\textbf{Non-conflict Cases.}\\
Do not judge the following as conflicts:
\begin{itemize}[leftmargin=*, itemsep=0pt, topsep=0pt]
    \item A low word limit with many content requirements, if the information is still compressible.
    \item Fixed opening/ending phrases with paragraph requirements, if they can appear at the start of the first paragraph and end of the last paragraph.
\end{itemize}

\textbf{Output Format.}\\
The output must contain two parts in order: \textbf{1) analysis:} briefly explain the checked constraints and whether they conflict in 1--5 sentences. \textbf{2) final:} output only \texttt{[[0]]} for conflict or \texttt{[[1]]} for no conflict.
\\

\textbf{Additional Requirement for \texttt{[[0]]}.}\\
If the output is \texttt{[[0]]}, the analysis must explicitly state which constraints are mutually exclusive and why they cannot be satisfied simultaneously. If the analysis only shows difficulty, tightness, or likely poor completion without identifying mutually exclusive constraints, the output must be \texttt{[[1]]}.
\\

\textbf{Few-shot Example.}\\
The few-shot examples are inserted from Table~\ref{tab:conflict-checker-few-shot}.
\\

\textbf{Real Input Instruction.}\\
Now process the real input. Determine only whether it contains a constraint conflict, without rewriting it.
\\

\textbf{Real input:} \texttt{\{input\}}
\\

Strictly follow the required output format: first provide \texttt{analysis}, then provide \texttt{final}.

\hrule

\end{minipage}
\end{table}

\begin{table}[htbp]
\centering
\caption{Few-shot examples for Filter filtering.}
\label{tab:conflict-checker-few-shot}

\begin{minipage}{1.0\textwidth}

\hrule

\textbf{Few-shot Example 1: Conflict.}
\textbf{Input:} Generate three paragraphs separated by \texttt{***}; the entire text must be lowercase; the second paragraph must begin with \texttt{Agreement}.
\textbf{Output:}
\texttt{analysis:}
The requirement that the entire text be lowercase conflicts with the requirement that the second paragraph begin with ``Agreement'', which requires an uppercase initial letter. These two constraints are mutually exclusive.
\texttt{final:}
\texttt{[[0]]}
\\
\\
\textbf{Few-shot Example 2: Conflict.}
\textbf{Input:} The output must not contain any punctuation; use commas to separate three entity names.
\textbf{Output:}
\texttt{analysis:}
The prohibition on punctuation disallows commas. However, the instruction requires commas to separate entities, which creates a direct conflict.
\texttt{final:}
\texttt{[[0]]}
\\
\\
\textbf{Few-shot Example 3: No Conflict.}
\textbf{Input:} Rewrite the sentence in passive voice; divide it into two sentences; the total length must be exactly 30 English words; use a formal academic tone.
\textbf{Output:}
\texttt{analysis:}
Passive voice, two-sentence structure, formal academic tone, and the 30-word constraint can be satisfied simultaneously through careful wording. There is no mutually exclusive constraint.
\texttt{final:}
\texttt{[[1]]}
\\
\\
\textbf{Few-shot Example 4: No Conflict.}
\textbf{Input:} Output two bullet points, each beginning with ``-''; do not use commas; include the phrase ``data science'' at least once.
\textbf{Output:}
\texttt{analysis:}
Two Markdown bullet points can be written without using commas. The phrase requirement and the comma prohibition are not contradictory.
\texttt{final:}
\texttt{[[1]]}
\\
\\
\textbf{Few-shot Example 5: No Conflict.}
\textbf{Input:} Write a book review in two paragraphs; the total length must not exceed 100 words; include at least three plot points and their effects; analyze the development of at least two main characters; begin with the phrase ``in the dystopian world of''; end with ``reflect on''; do not mention film adaptations or sequels.
\textbf{Output:}
\texttt{analysis:}
Although the word limit is tight and there are many content requirements, no mutually exclusive constraints exist. A highly concise summary could still satisfy all requirements.
\texttt{final:}
\texttt{[[1]]}

\hrule

\end{minipage}
\end{table}

\begin{table}[htbp]
\centering
\caption{Prompt for constraint satisfaction checking of Judger.}
\label{tab:constraint-checking-query-template}

\begin{minipage}{1.0\textwidth}

\hrule

Please judge whether the given reply follows the constraint(s). Analyze each constraint one by one and determine if it is satisfied.

\medskip

The few-shot examples are inserted from Table~\ref{tab:few-shot-constraint-checking}.

\medskip

Now judge the following:

\medskip

\texttt{[Reply]}

\texttt{\{resp\_to\_check\}}

\medskip

\texttt{[Constraint]}

\texttt{\{con\}}

\medskip

Output your analysis and then the final score in \texttt{[[score]]} format. If all constraints are satisfied, output \texttt{[[1]]}; otherwise, output \texttt{[[0]]}.
\hrule

\end{minipage}
\end{table}

\begin{table}[htbp]
\centering
\caption{Few-shot examples for constraint satisfaction checking of Judger.}
\label{tab:few-shot-constraint-checking}

\begin{minipage}{1.0\textwidth}

\hrule

\textbf{Example 1.}
\texttt{[Reply]} Kathy and Sue are the two characters in this story.
\texttt{[Constraint]} The response should include at least three characters from the story.
\texttt{[Analysis]} The reply mentions only ``Kathy'' and ``Sue'', which is 2 characters. The constraint requires at least 3 characters.
NOT SATISFIED $\rightarrow$ \texttt{[[0]]}
\\
\\
\textbf{Example 2.}
\texttt{[Reply]} Kathy\par
Sue\par
John
\texttt{[Constraint]} The response should include at least three characters from the story.
\texttt{[Analysis]} The reply mentions ``Kathy'', ``Sue'', and ``John'', which is 3 characters. This satisfies the constraint.
SATISFIED $\rightarrow$ \texttt{[[1]]}
\\
\\
\textbf{Example 3.}
\texttt{[Reply]} The characters in this story are Kathy and Sue.\par
Kathy is mentioned in multiple sentences.\par
Sue appears in the conflict described.
\texttt{[Constraint]} Each character's name must appear in a different paragraph.
\texttt{[Analysis]} The reply has 3 paragraphs: 1) ``The characters in this story are Kathy and Sue.'' 2) ``Kathy is mentioned...'' 3) ``Sue appears...'' Each paragraph contains a character name.
SATISFIED $\rightarrow$ \texttt{[[1]]}
\\
\\
\textbf{Example 4.}
\texttt{[Reply]} Kathy and Sue are friends. They talk every day.
\texttt{[Constraint]} Each character's name must appear in a different paragraph.
\texttt{[Analysis]} The reply has only 1 paragraph. Both character names appear in the same paragraph.
NOT SATISFIED $\rightarrow$ \texttt{[[0]]}
\\
\\
\textbf{Example 5.}
\texttt{[Reply]} The answer is five.
\texttt{[Constraint]} The response must include the word `count' exactly once.
\texttt{[Analysis]} The reply does not contain the word ``count''.
NOT SATISFIED $\rightarrow$ \texttt{[[0]]}
\\
\\
\textbf{Example 6.}
\texttt{[Reply]} Let me count the items for you.
\texttt{[Constraint]} The response must include the word `count' exactly once.
\texttt{[Analysis]} The reply contains ``count'' exactly once in ``Let me count''. No other occurrences of ``count''.
SATISFIED $\rightarrow$ \texttt{[[1]]}
\\
\\
\textbf{Example 7.}
\texttt{[Reply]} I will count the number of items. Please count them carefully.
\texttt{[Constraint]} The response must include the word `count' exactly once.
\texttt{[Analysis]} The reply contains ``count'' twice: ``I will count'' and ``Please count them''. This is more than once.
NOT SATISFIED $\rightarrow$ \texttt{[[0]]}
\\
\\
\textbf{Example 8.}
\texttt{[Reply]} Kathy Sue John
\texttt{[Constraint]} The response should not use any commas.
\texttt{[Analysis]} The reply contains no commas, only spaces between the names.
SATISFIED $\rightarrow$ \texttt{[[1]]}
\\
\\
\textbf{Example 9.}
\texttt{[Reply]} Kathy, Sue, and John are the characters.
\texttt{[Constraint]} The response should not use any commas.
\texttt{[Analysis]} The reply contains commas: ``Kathy, Sue, and John''.
NOT SATISFIED $\rightarrow$ \texttt{[[0]]}

\hrule

\end{minipage}
\end{table}

\begin{algorithm}[htbp]
\caption{Self-Evolving Instruction Following}
\label{alg:seif}
\begin{algorithmic}[1]
\Require Seed instruction set $\mathcal{D}_{seed}$; initial Instructor $I_{\psi_0}$; initial Follower $F_{\theta_0}$; number of iterations $T$; group size $G$
\Ensure Trained Instructor $I_{\psi_T}$ and Follower $F_{\theta_T}$

\For{$t = 0,1,\dots,T-1$}

    \State Instantiate frozen Filter $Q_t \leftarrow F_{\theta_t}$
    \State Instantiate frozen Judger $J_t \leftarrow F_{\theta_t}$

    \Statex
    \Comment{\textbf{Instructor optimization}}
    \State Freeze $F_{\theta_t}$, $Q_t$, and $J_t$
    \State Initialize working Instructor $I_{\psi} \leftarrow I_{\psi_t}$

    \For{each seed instruction $z \sim \mathcal{D}_{seed}$}
        \State Sample a group of evolved instructions 
        $\{x_i\}_{i=1}^{G} \sim I_{\psi}(\cdot \mid z)$

        \For{$i = 1,\dots,G$}
            \If{$Q_t(x_i)=0$}
                \State Assign Instructor reward $r_i^I \leftarrow 0$
            \Else
                \State Sample response $y_i \sim F_{\theta_t}(\cdot \mid x_i)$
                \State Compute satisfaction score 
                $s_i \leftarrow J_t(x_i,y_i)$
                \State Assign Instructor reward 
                $r_i^I \leftarrow 1 - s_i$
            \EndIf
        \EndFor

        \If{$\mathrm{std}(\{r_i^I\}_{i=1}^{G}) = 0$}
            \State Skip this Instructor update
        \Else
            \State Compute group-relative advantages 
            $\{A_i^I\}_{i=1}^{G}$ from rewards $\{r_i^I\}_{i=1}^{G}$
            \State Update working Instructor $I_{\psi}$ by GRPO using 
            $\{(z,x_i,A_i^I)\}_{i=1}^{G}$
        \EndIf
    \EndFor

    \State Set updated Instructor $I_{\psi_{t+1}} \leftarrow I_{\psi}$

    \Statex
    \Comment{\textbf{Follower optimization}}
    \State Instantiate frozen Judger $J'_t \leftarrow F_{\theta_t}$
    \State Freeze $I_{\psi_{t+1}}$, $Q_t$, and $J'_t$
    \State Initialize working Follower $F_{\theta} \leftarrow F_{\theta_t}$

    \For{each seed instruction $z \sim \mathcal{D}_{seed}$}
        \State Generate evolved instruction 
        $x \sim I_{\psi_{t+1}}(\cdot \mid z)$

        \If{$Q_t(x)=0$}
            \State Skip $x$
        \Else
            \State Sample a group of responses 
            $\{y_i\}_{i=1}^{G} \sim F_{\theta}(\cdot \mid x)$

            \For{$i = 1,\dots,G$}
                \State Compute Follower reward 
                $r_i^F \leftarrow J'_t(x,y_i)$
            \EndFor

            \If{$\mathrm{std}(\{r_i^F\}_{i=1}^{G}) = 0$}
                \State Skip this Follower update
            \Else
                \State Compute group-relative advantages 
                $\{A_i^F\}_{i=1}^{G}$ from rewards $\{r_i^F\}_{i=1}^{G}$
                \State Update working Follower $F_{\theta}$ by GRPO using 
                $\{(x,y_i,A_i^F)\}_{i=1}^{G}$
            \EndIf
        \EndIf
    \EndFor

    \State Set updated Follower $F_{\theta_{t+1}} \leftarrow F_{\theta}$

\EndFor

\State \Return $I_{\psi_T}, F_{\theta_T}$
\end{algorithmic}
\end{algorithm}

\section{Benchmark Details}
\label{sec:benchmark}

We evaluate SEIF on six widely adopted instruction-following benchmarks, each covering a distinct dimension of instruction-following capability.

\textbf{IFEval} is a foundational benchmark that introduces verifiable constraints---objectively checkable requirements such as ``write in more than 400 words'' or ``include keyword $X$ exactly three times''---into natural language instructions. Responses are automatically evaluated via constraint-checking scripts, enabling reproducible assessment without human annotation. It consists of approximately 500 prompts with 25 constraint templates.

\textbf{CFBench} evaluates LLMs on their ability to follow complex, multi-constraint instructions across more than 200 real-life scenarios and 50 NLP tasks. It features a systematic taxonomy of 10 primary constraint categories and over 25 subcategories, with Chinese-language instructions comprising a significant portion of the dataset. Evaluation integrates multi-dimensional criteria aligned with user perceptions of constraint fulfillment.

\textbf{FollowBench} introduces a multi-level fine-grained evaluation framework for instruction following. It categorizes constraints into five types (content, situation, style, format, and example) and progressively increases instruction difficulty by incrementally adding constraints at each level. Key metrics include Hard Satisfaction Rate (HSR), Soft Satisfaction Rate (SSR), and Consistent Satisfaction Levels (CSL).

\textbf{WritingBench} is a comprehensive writing-focused benchmark spanning 6 core domains (Academic, Finance, Politics, Literature, Education, and Advertising) and 100 subdomains. Each of its 1,000 writing queries includes an average of 1,500+ tokens, and evaluation employs a fine-tuned critic model that generates 5 instance-specific criteria for criteria-aware scoring across style, format, and length dimensions.

\textbf{AgentIF} is the first benchmark specifically designed for instruction following in agentic scenarios. It contains 707 human-annotated instructions from 50 real-world agentic applications, with each instruction averaging 1,723 words and 11.9 constraints covering tool specifications, condition constraints, formatting, and safety requirements. Evaluation combines code-based, LLM-based, and hybrid code-LLM approaches.

\textbf{Multi-IF} extends instruction-following evaluation to multi-turn and multilingual settings. It comprises 4,501 multilingual conversations with three turns each across 8 languages (English plus 7 translations), revealing that all state-of-the-art models exhibit higher failure rates with additional conversation turns, with non-Latin-script languages showing particularly elevated error rates.

\section{Baseline Details}
\label{sec:baseline}
We compare SEIF against baselines covering specialized instruction-following models and self-training methods.

\subsection{Specialized Instruction-Following Models}

\textbf{Self-Supervised-7B} improves instruction following via label-free self-supervised RL. It decomposes multi-constraint instructions into incremental curricula, derives pseudo-labels from the instructions themselves to train constraint-wise reward models, and combines rule-based verification for hard constraints with a self-supervised classifier for soft constraints. The resulting 7B model achieves consistent gains on in-domain instruction-following benchmarks, including improvements of +5.0 on IFEval, +5.0 on CFBench, and +2.4 on FollowBench over Qwen2.5-7B-Instruct.

\textbf{VERIF-8B} applies verification engineering to RL-based instruction-following training. It combines rule-based code verification for hard constraints (format, keywords, length) with LLM-based reasoning verification (using QwQ-32B) for soft constraints (style, semantic requirements), trained on the VerInstruct dataset of approximately 22,000 instruction-following instances. It achieves state-of-the-art performance among 8B-scale models.

\textbf{RAIF-7B} improves instruction following through a two-stage RL approach that decomposes complex instructions and trains verifiable rule-centric reward signals. It incorporates sample-wise contrast to enforce authentic chain-of-thought reasoning and behavior cloning of expert reasoners, achieving significant gains even for 1.5B models (11.74\% improvement on IFEval).

\textbf{SPAR-8B-DPO} combines iterative self-play with tree-search-based refinement to generate high-quality preference pairs for DPO training. Using breadth-first and depth-first search strategies, it produces preference pairs free from interfering content variations. A LLaMA3-8B model trained through three SPAR iterations surpasses GPT-4-Turbo on IFEval (81.3\%).

\textbf{Crab-7B-DPO} uses a novel constraint back-translation technique that extracts implicit constraints already satisfied by high-quality responses and re-injects them into instructions. This reduces data noise and generation cost compared to generating complex instructions from scratch. 

\textbf{Conifer-7B-DPO} curates a high-quality instruction-tuning dataset using multi-round GPT-4-driven refinement from ShareGPT seed data. It employs a progressive easy-to-hard learning scheme with process feedback, enabling a 7B model to outperform open-source models 10x larger on complex constrained instruction-following tasks.

\subsection{Baseline Method}

When reproducing these baselines, we first use the backbone model, following our experimental settings, with the prompt in Table~\ref{tab:prompt-soft} to add either soft or hard constraints to the collected 5,120 seed instructions, forming the initial dataset. We then follow the remaining settings of the original papers for reproduction. 

\subsubsection{SFT and ProxyReward}

\textbf{SFT}  We use GPT-5.1 to generate responses to complex instructions of the initial dataset and perform full-parameter SFT with LLaMA-Factory~\citep{zheng2024llamafactory} on 8 H200 GPUs. \textbf{ProxyReward} introduces a reinforcement-learning framework for open-ended long-context generation, automatically constructing proxy question-answer pairs and using targeted reward signals to evaluate the comprehensiveness and accuracy of generated responses, thereby improving long-form generation without relying on gold-standard reference answers.

\subsubsection{Self-Play Baselines}

\textbf{Self-Correct}  enables models to self-identify and correct instruction-following errors through iterative refinement, using the model's own judgment to improve response quality without external supervision.

\textbf{Humpback}  applies instruction back-translation to unlabeled web data: a seed model generates instruction-response pairs from web content, then self-curates high-quality examples through iterative improvement without distillation from larger models.

\textbf{SELF}  fine-tunes models on their own high-quality outputs in an iterative self-training loop, progressively improving instruction-following capability through self-generated data.

\textbf{Self-Rewarding} uses the LLM itself as a reward model via LLM-as-a-Judge prompting, combined with iterative DPO training. This allows models to simultaneously improve both instruction-following ability and reward-generation capability. Fine-tuned Llama 2 70B with three iterations outperforms Claude 2, Gemini Pro, and GPT-4 0613 on AlpacaEval 2.0.

\textbf{I-SHEEP} enables continuous self-alignment from scratch through a cycle of self-synthesis, self-assessment, filtering, and supervised fine-tuning. On Qwen-1.5 72B, it achieves 8.88\% absolute improvement on IFEval and 78.2\% relative improvement on AlpacaEval.

\textbf{Meta-Rewarding} extends self-rewarding by adding a meta-layer where models judge their own judgments to refine judgment quality. It achieves unsupervised improvement of Llama-3-8B-Instruct from 22.9\% to 39.4\% win rate on AlpacaEval 2.0 without human supervision.

\section{Judger and Filter Reliability}
\label{sec:H}
\begin{figure}[htbp]
    \centering
    \small
    \begin{minipage}[t]{0.46\textwidth}
        \centering
        \setlength{\tabcolsep}{1.25mm}
        \renewcommand{\arraystretch}{1.05}
        \captionof{table}{Correlation between Filter models from different iteration turns and human judgments on conflicting-instruction filtering.}
        \label{tab:filter_results}
        \begin{tabular}{lcc}
            \toprule[1pt]
            \textbf{Model} & \textbf{Accuracy} & \textbf{F1} \\
            \midrule
            Filter-T1 & 0.80 & 0.80 \\
            Filter-T2 & 0.80 & 0.79 \\
            Filter-T3 & 0.79 & 0.78 \\
            \bottomrule[1pt]
        \end{tabular}
    \end{minipage}
    \hspace{0.02\textwidth}
    \begin{minipage}[t]{0.46\textwidth}
        \centering
        \setlength{\tabcolsep}{1.25mm}
        \renewcommand{\arraystretch}{1.05}
        \captionof{table}{Correlation between Judger models from different iteration turns and human judgments in evaluating constraint following.}
        \label{tab:judger_results}
        \begin{tabular}{lcc}
            \toprule[1pt]
            \textbf{Model} & \textbf{Accuracy} & \textbf{F1} \\
            \midrule
            Judger-T1 & 0.73 & 0.71 \\
            Judger-T2 & 0.74 & 0.72 \\
            Judger-T3 & 0.73 & 0.70 \\
            \bottomrule[1pt]
        \end{tabular}
    \end{minipage}
\end{figure}

We evaluate the reliability of the self-instantiated Filter and Judger from three complementary perspectives. First, we measure their agreement with human judgments across three self-evolution turns. Second, we ablate the initialization source of the Filter and Judger to examine whether refreshing them from the latest Follower is beneficial. Third, we conduct a human evaluation on final model outputs to verify that the improvements are not merely artifacts of the self-instantiated Judger.

\paragraph{Agreement with human judgments.}
We first sample a 400-example evaluation set from VerInstruct~\citep{peng2025verif}. 
For Judger evaluation, we use Qwen2.5-7B-Instruct to generate responses to these instructions, and human annotators manually annotate whether each constraint is satisfied. 
For Filter evaluation, human annotators manually modify the constraints so that half of the instructions contain conflicting constraints, and then evaluate whether the Filter can identify such conflicts. 
As shown in Table~\ref{tab:filter_results}, the Filter achieves stable performance across Turn 1 to Turn 3, with Accuracy ranging from 0.79 to 0.80 and F1 ranging from 0.78 to 0.80. 
These results indicate that the self-instantiated Filter can consistently identify instructions with conflicting constraints throughout the iterative self-evolution process.  As shown in Table~\ref{tab:judger_results}, the Judger obtains Accuracy ranging from 0.73 to 0.74 and F1 ranging from 0.70 to 0.72.  Although judging constraint satisfaction is more challenging and subjective than detecting explicit conflicts, the Judger still shows reasonable agreement with human annotations. 
Notably, both modules exhibit only small fluctuations across turns, suggesting that self-instantiation from the evolving Follower does not lead to obvious reliability degradation.  Details of the human annotation protocol are provided in Appendix~\ref{sec:Crowdsourcing}.

\paragraph{Effect of refreshing the Filter and Judger.}
To further examine whether the adaptive Filter and Judger are useful, we conduct an ablation study on Qwen2.5-7B-Instruct by varying whether the Filter and Judger are initialized from the base model or refreshed from the latest Follower. As shown in Table~\ref{tab:filter_judger_source_ablation}, using both the latest Filter and latest Judger achieves the best performance on all three benchmarks. Compared with using both modules initialized from the base model, refreshing both modules from the latest Follower improves IFEval from 76.8 to 78.6, CFBench from 49.0 to 51.0, and FollowBench from 57.1 to 59.0. These results suggest that adaptive filtering and adaptive reward evaluation are both important for maintaining useful supervision during self-evolution.

\begin{table}[t]
\centering
\caption{Ablation study on the initialization source of the Filter and Judger using SEIF-7B.}
\label{tab:filter_judger_source_ablation}
\resizebox{\textwidth}{!}{
\begin{tabular}{llccc}
\toprule
Filter & Judger & IFEval & CFBench & FollowBench \\
\midrule
initialized from the base model & initialized from the base model & 76.8 & 49.0 & 57.1 \\
latest Follower & initialized from the base model & 77.5 & 49.0 & 57.8 \\
initialized from the base model & latest Follower & 77.7 & 50.0 & 58.4 \\
latest Follower & latest Follower & \textbf{78.6} & \textbf{51.0} & \textbf{59.0} \\
\bottomrule
\end{tabular}
}
\end{table}

\paragraph{Human evaluation on final outputs.}
While the above agreement analysis evaluates whether the Filter and Judger align with human judgments, it does not directly show whether the final trained model produces responses that humans prefer. Therefore, we further conduct a pairwise human evaluation on final model outputs. For each sampled instruction, we collect responses from BASE, Ours-SEIF, w/o Instructor Evolving, and Meta-Rewarding on Qwen2.5-7B-Instruct.  Given the same instruction and two anonymized responses, annotators are asked to choose which response better follows the instruction, with three possible labels: A better, B better, or Tie. Annotators are instructed to prioritize the constraint satisfaction rate. If both responses satisfy the same proportion of constraints, they then compare task relevance, completeness, helpfulness, fluency, and overall quality. A tie is selected when both responses are similarly good, similarly flawed, or when the difference is too minor to determine a clear preference. As shown in Table~\ref{tab:human_pairwise_eval}, SEIF is preferred over BASE in 62.8\% of cases, with only 19.7\% losses. SEIF also outperforms w/o Instructor Evolving and Meta-Rewarding, achieving win rates of 56.5\% and 53.5\%, respectively. These results provide direct human-evaluation evidence that SEIF improves final instruction-following quality, rather than merely optimizing for the bias of the  Judger.

\begin{table}[t]
\centering
\caption{Pairwise human evaluation on sampled instructions from VerInstruct. Results are reported as percentages.}
\label{tab:human_pairwise_eval}
\begin{tabular}{lccc}
\toprule
Comparison & SEIF Win & Tie & SEIF Lose \\
\midrule
SEIF vs BASE & \textbf{62.8} & 17.5 & 19.7 \\
SEIF vs w/o Instructor Evolving & \textbf{56.5} & 21.5 & 22.0 \\
SEIF vs Meta-Rewarding & \textbf{53.5} & 21.0 & 25.5 \\
\bottomrule
\end{tabular}
\end{table}

\section{Instruction Evolving Example}
\label{sec:iee}
We provide examples showing how our instructions evolve across training iterations below.

\definecolor{lightblue}{RGB}{230, 242, 255}
\definecolor{lightyellow}{RGB}{255, 253, 208}
\definecolor{lightgreen}{RGB}{220, 252, 220}
\definecolor{headerbg}{RGB}{44, 62, 80}
\definecolor{row1bg}{RGB}{247, 250, 252}
\definecolor{row2bg}{RGB}{255, 255, 255}

\subsection{Case 1: Restaurant Description -- Midsummer House}

\textbf{Task:} \texttt{Convert restaurant key-value pairs into a fluent English sentence. input: name[Midsummer House], food[Indian], customer rating[low], near[Café Rouge]}

\begin{longtable}{|p{0.22\linewidth}|p{2.4in}|p{1.3in}|}
\hline
\rowcolor{headerbg}
\textcolor{white}{\textbf{Round}} & \textcolor{white}{\textbf{Constraints (in order)}} & \textcolor{white}{\textbf{Type}} \\
\hline
\endfirsthead
\hline
\rowcolor{headerbg}
\textcolor{white}{\textbf{Round}} & \textcolor{white}{\textbf{Constraints}} & \textcolor{white}{\textbf{Type}} \\
\hline
\endhead
\endfoot

\rowcolor{lightblue}
\textbf{T1 ($N=3$)} &
\begin{enumerate}[left=0pt, nosep, label={[\theenumi]}]
  \item Include the \textbf{name}, \textbf{type of food}, \textbf{customer rating}, and \textbf{nearby location} in a single sentence.
  \item Word Count: at least \textbf{20 words}.
  \item Highlight the \textbf{name} and the \textbf{nearby location} using markdown emphasis.
\end{enumerate} &
Information coverage \newline Word count \newline Format/highlight \\
\hline

\rowcolor{lightyellow}
\textbf{T2 ($N=3$)} &
\begin{enumerate}[left=0pt, nosep, label={[\theenumi]}]
  \item Include the \textbf{name}, \textbf{type of food}, \textbf{customer rating}, and \textbf{nearby restaurant} in a single sentence.
  \item Word Count: at least \textbf{25 words}.
  \item Highlight the \textbf{name} and the \textbf{nearby restaurant} using bold markdown.
\end{enumerate} &
Information coverage \newline Higher word count \newline Format/bold \\
\hline

\rowcolor{lightgreen}
\textbf{T3 ($N=5$)} &
\begin{enumerate}[left=0pt, nosep, label={[\theenumi]}]
  \item Use specific adjectives to describe the restaurant's \textbf{ambiance} and \textbf{service quality}.
  \item Limit the sentence to \textbf{30 words}.
  \item Mention the nearby restaurant \textbf{`Café Rouge'} in the sentence.
  \item Write the sentence in the \textbf{past tense} to reflect historical customer experiences.
  \item Include a clause about the type of \textbf{Indian cuisine} offered.
\end{enumerate} &
Descriptive specificity \newline Word limit \newline Lexical content \newline Tense constraint \newline Cuisine clause \\
\hline
\end{longtable}

\textbf{Analysis:} T1 requires basic field coverage and markdown emphasis; T2 raises the minimum word count and changes the highlighted entity from nearby location to nearby restaurant with bold formatting; T3 adds descriptive adjectives, past tense, an upper word limit, and a cuisine-specific clause, making the sentence harder to control.

\newpage
\subsection{Case 2: Electric Vehicle Conversation Generation}

\textbf{Task:} \texttt{Construct a conversation between two people related to the topic. input: Driving electric vehicles}

\begin{longtable}{|p{0.22\linewidth}|p{2.4in}|p{1.3in}|}
\hline
\rowcolor{headerbg}
\textcolor{white}{\textbf{Round}} & \textcolor{white}{\textbf{Constraints}} & \textcolor{white}{\textbf{Type}} \\
\hline

\rowcolor{lightblue}
\textbf{T1 ($N=3$)} &
\begin{enumerate}[left=0pt, nosep, label={[\theenumi]}]
  \item The response should include at least \textbf{5 sentences} about the advantages and disadvantages of driving electric vehicles.
  \item Highlight the \textbf{environmental impact} and \textbf{cost benefits} of electric vehicles using markdown emphasis.
  \item Your answer must contain \textbf{[electricity]}, \textbf{[battery]}, and \textbf{[range]} as placeholders, and be written entirely in \textbf{English}, avoiding \textbf{ALL CAPS}.
\end{enumerate} &
Sentence count \newline Topic coverage \newline Markdown emphasis \newline Placeholder inclusion \newline English language \newline Case restriction \\
\hline

\rowcolor{lightyellow}
\textbf{T2 ($N=3$)} &
\begin{enumerate}[left=0pt, nosep, label={[\theenumi]}]
  \item The response should include exactly \textbf{3 bullet points} using markdown bullets.
  \item The response must contain at least \textbf{2 highlighted spans}.
  \item Your answer must be in \textbf{English}, and in \textbf{all lowercase letters}; no capital letters allowed.
\end{enumerate} &
Bullet count \newline Markdown bullets \newline Highlighted spans \newline English language \newline Lowercase restriction \\
\hline

\rowcolor{lightgreen}
\textbf{T3 ($N=5$)} &
\begin{enumerate}[left=0pt, nosep, label={[\theenumi]}]
  \item Mandatory use of the terms \textbf{`sustainability'}, \textbf{`reduction'}, and \textbf{`emissions'} in the conversation.
  \item Limit the conversation to \textbf{three sentences per person}.
  \item Write the conversation in \textbf{Markdown format}, including headers for each speaker's lines.
  \item Ensure the conversation reflects a \textbf{pragmatic context}, focusing on the benefits and concerns of electric vehicles.
  \item Emulate the style of a \textbf{technology blogger} discussing the latest trends in eco-friendly transportation.
\end{enumerate} &
Mandatory terms \newline Sentence limit \newline Markdown format \newline Speaker headers \newline Pragmatic context \newline Style imitation \\
\hline
\end{longtable}

\textbf{Analysis:} T1 requires a multi-sentence English conversation covering both advantages and disadvantages of electric vehicles, with markdown emphasis and required placeholders; T2 shifts to a stricter bullet-list format with lowercase-only English and highlighted spans; T3 further increases complexity by requiring specific sustainability-related terms, speaker-level sentence limits, markdown speaker headers, a pragmatic discussion context, and a technology-blogger style.

\subsection{Case 3: Restaurant Review -- xname}

\textbf{Task:} \texttt{Write a brief restaurant review using all provided information. input: name[xname], recommend[yes], cuisine[Italian], decor[bad], location[the Upper East Side], price[expensive], service[acceptable]}

\begin{longtable}{|p{0.22\linewidth}|p{2.4in}|p{1.3in}|}
\hline
\rowcolor{headerbg}
\textcolor{white}{\textbf{Round}} & \textcolor{white}{\textbf{Constraints}} & \textcolor{white}{\textbf{Type}} \\
\hline

\rowcolor{lightblue}
\textbf{T1 ($N=3$)} &
\begin{enumerate}[left=0pt, nosep, label={[\theenumi]}]
  \item Response should include exactly \textbf{3 sentences}.
  \item Response must contain the word \textbf{`Italian'} at least \textbf{twice}.
  \item Response should not use the word \textbf{`bad'}.
\end{enumerate} &
Sentence count \newline Lexical frequency \newline Forbidden word \\
\hline

\rowcolor{lightyellow}
\textbf{T2 ($N=3$)} &
\begin{enumerate}[left=0pt, nosep, label={[\theenumi]}]
  \item Response should contain exactly \textbf{2 sentences}.
  \item Response should include the word \textbf{`expensive'} at least once.
  \item Highlight the restaurant name using \textbf{bold markdown}.
\end{enumerate} &
Sentence count \newline Lexical content \newline Format/bold \\
\hline

\rowcolor{lightgreen}
\textbf{T3 ($N=5$)} &
\begin{enumerate}[left=0pt, nosep, label={[\theenumi]}]
  \item Use \textbf{all of the information} provided.
  \item Write in an \textbf{encyclopedic style}.
  \item Limit to \textbf{two sentences}.
  \item Write with a \textbf{polite tone}.
  \item Mandate the use of the phrase \textbf{`worth the try'} if recommending.
\end{enumerate} &
Complete coverage \newline Register/style \newline Sentence limit \newline Tone constraint \newline Mandatory phrase \\
\hline
\end{longtable}

\textbf{Analysis:} T1 controls sentence count and avoids a direct negative adjective; T2 reduces the sentence count and adds bold formatting; T3 requires full information coverage, encyclopedic style, politeness, and the exact recommendation phrase `worth the try', which makes the review more constrained semantically and stylistically.

\subsection{Case 4: Navy News Article Summary}

\textbf{Task:} \texttt{Generate a short highlight summary from a Navy news article about ending all-caps messaging and adopting NICE.}

\begin{longtable}{|p{0.22\linewidth}|p{2.4in}|p{1.3in}|}
\hline
\rowcolor{headerbg}
\textcolor{white}{\textbf{Round}} & \textcolor{white}{\textbf{Constraints}} & \textcolor{white}{\textbf{Type}} \\
\hline

\rowcolor{lightblue}
\textbf{T1 ($N=3$)} &
\begin{enumerate}[left=0pt, nosep, label={[\theenumi]}]
  \item Response should contain exactly \textbf{8 lines}.
  \item Highlight at least \textbf{3 sections} in the summary using bold text.
  \item Answer must end with the exact phrase: \textbf{`The Navy is evolving.'}
\end{enumerate} &
Line count \newline Format/highlight \newline Exact ending phrase \\
\hline

\rowcolor{lightyellow}
\textbf{T2 ($N=3$)} &
\begin{enumerate}[left=0pt, nosep, label={[\theenumi]}]
  \item Highlight key points using exactly \textbf{5 bullet points}.
  \item Highlight spans of text by wrapping them in asterisks \texttt{*like this*}.
  \item Response must contain \textbf{no commas} in any part of the text.
\end{enumerate} &
Bullet count \newline Highlight format \newline Forbidden punctuation \\
\hline

\rowcolor{lightgreen}
\textbf{T3 ($N=5$)} &
\begin{enumerate}[left=0pt, nosep, label={[\theenumi]}]
  \item Mandatory use of the term \textbf{`all caps'} and explain its impact on readability.
  \item Include at least one example of how the new system, \textbf{NICE}, is more efficient.
  \item Limit the response to exactly \textbf{10 lines}.
  \item Use a \textbf{formal tone} in the summary.
  \item Highlight the \textbf{cost savings} and the \textbf{switch to the new system} in two sentences.
\end{enumerate} &
Meta-level lexical \newline Semantic example \newline Line precision \newline Formal register \newline Content allocation \\
\hline
\end{longtable}

\textbf{Analysis:} T1 is line-count and ending-phrase oriented; T2 changes the output into exactly five highlighted bullet points while banning commas; T3 adds a meta-level explanation of `all caps', requires an efficiency example for NICE, fixes the output at ten lines, and allocates highlighted content across two sentences.

\subsection{Case 5: WWF Five-Sentence Summary}

\textbf{Task:} \texttt{Summarize the given text using five sentences. input: A passage about the World Wildlife Fund, its conservation mission, ecosystems, wildlife, and global environmental protection.}

\begin{longtable}{|p{0.22\linewidth}|p{2.4in}|p{1.3in}|}
\hline
\rowcolor{headerbg}
\textcolor{white}{\textbf{Round}} & \textcolor{white}{\textbf{Constraints}} & \textcolor{white}{\textbf{Type}} \\
\hline
\endfirsthead
\hline
\rowcolor{headerbg}
\textcolor{white}{\textbf{Round}} & \textcolor{white}{\textbf{Constraints}} & \textcolor{white}{\textbf{Type}} \\
\hline
\endhead
\endfoot

\rowcolor{lightblue}
\textbf{T1 ($N=3$)} &
\begin{enumerate}[left=0pt, nosep, label={[\theenumi]}]
  \item The response should include exactly three required keywords: \textbf{World Wildlife Fund}, \textbf{conservation}, and \textbf{ecosystems}.
  \item The response must contain exactly two paragraphs separated by \texttt{***}.
  \item The response should have a word count of less than \textbf{100 words}.
\end{enumerate} &
Keyword inclusion \newline Paragraph structure \newline Word limit \\
\hline

\rowcolor{lightyellow}
\textbf{T2 ($N=5$)} &
\begin{enumerate}[left=0pt, nosep, label={[\theenumi]}]
  \item Use specific conservation terms such as \textbf{ecosystems}, \textbf{wildlife}, and \textbf{biodiversity} in the summary.
  \item Limit the summary to exactly \textbf{five sentences}.
  \item Write in an \textbf{informative and factual tone}.
  \item Exclude any mention of WWF's \textbf{establishment date}.
  \item Emphasize the impact of WWF's work on \textbf{global environmental protection}.
\end{enumerate} &
Domain terminology \newline Sentence count \newline Register constraint \newline Exclusion constraint \newline Impact emphasis \\
\hline

\rowcolor{lightgreen}
\textbf{T3 ($N=5$)} &
\begin{enumerate}[left=0pt, nosep, label={[\theenumi]}]
  \item Use specific terms such as \textbf{conservation}, \textbf{ecosystems}, and \textbf{International Union for Conservation of Nature}.
  \item Limit the sentence count to exactly \textbf{five sentences}.
  \item Write in \textbf{Markdown format}, with appropriate headers and emphasis for key points.
  \item Adopt a \textbf{formal and informative tone}.
  \item Include a comparison between WWF's current state and its establishment in \textbf{1961}.
\end{enumerate} &
Specific terminology \newline Sentence precision \newline Markdown formatting \newline Formal register \newline Historical comparison \\
\hline
\end{longtable}

\textbf{Analysis:} T1 imposes basic keyword coverage, paragraph separation, and a word limit; T2 increases difficulty by requiring exact sentence count, conservation-specific terminology, factual tone, exclusion of the establishment date, and global impact emphasis; T3 further strengthens the task by requiring Markdown structure, formal presentation, exact five-sentence control, and a comparison between WWF's current role and its origin in 1961.

\subsection{Case 6: Gangs of New York Question Generation}

\textbf{Task:} \texttt{Generate a relevant question about interpretations of the phrase ``The blood stays on the blade'' from a movie discussion.}

\begin{longtable}{|p{0.22\linewidth}|p{2.4in}|p{1.3in}|}
\hline
\rowcolor{headerbg}
\textcolor{white}{\textbf{Round}} & \textcolor{white}{\textbf{Constraints}} & \textcolor{white}{\textbf{Type}} \\
\hline

\rowcolor{lightblue}
\textbf{T1 ($N=5$)} &
\begin{enumerate}[left=0pt, nosep, label={[\theenumi]}]
  \item Mandatory use of the phrase \textbf{`overtly religious themes'} in the response.
  \item Include at least one reference to a specific \textbf{movie character} in the question.
  \item Write the question in a \textbf{rhetorical style}.
  \item Limit the question to \textbf{20 words}.
  \item Output the question as a \textbf{bullet point}.
\end{enumerate} &
Mandatory phrase \newline Character reference \newline Rhetorical style \newline Word limit \newline Bullet format \\
\hline

\rowcolor{lightyellow}
\textbf{T2 ($N=5$)} &
\begin{enumerate}[left=0pt, nosep, label={[\theenumi]}]
  \item Use the phrase \textbf{`The blood stays on the blade'} in the question.
  \item Limit the question to \textbf{50 words}.
  \item Write the question in an \textbf{interrogative form} that reflects religious themes.
  \item Include at least one \textbf{example from the movie} to illustrate the question.
  \item Output the question in \textbf{Markdown format}.
\end{enumerate} &
Exact phrase \newline Word limit \newline Interrogative style \newline Example inclusion \newline Markdown format \\
\hline

\rowcolor{lightgreen}
\textbf{T3 ($N=5$)} &
\begin{enumerate}[left=0pt, nosep, label={[\theenumi]}]
  \item Use at least \textbf{three different interpretations} of the phrase in the question.
  \item Limit the question to \textbf{40 words}.
  \item Emulate the tone of a \textbf{film critic}.
  \item Include a \textbf{rhetorical question} to engage the reader.
  \item Write the question in Markdown format, using \textbf{bold} and \textbf{italics} for emphasis.
\end{enumerate} &
Semantic multiplicity \newline Word limit \newline Style imitation \newline Rhetorical device \newline Markdown emphasis \\
\hline
\end{longtable}

\textbf{Analysis:} T1 requires a compact rhetorical bullet with a character reference and a fixed thematic phrase; T2 embeds the exact quote and a movie example in markdown; T3 raises semantic density by requiring three interpretations while also demanding a critic-like tone, rhetorical engagement, and both bold and italic markdown.

\subsection{Case 7: Employee Monthly Performance Rating}

\textbf{Task:} \texttt{On a scale of 1 to 5, rate the employee's performance in the past month. input: Assisted with 3 customer service inquiries; provided technical support to 2 customers; resolved 3 project-related issues.}

\begin{longtable}{|p{0.22\linewidth}|p{2.4in}|p{1.3in}|}
\hline
\rowcolor{headerbg}
\textcolor{white}{\textbf{Round}} & \textcolor{white}{\textbf{Constraints}} & \textcolor{white}{\textbf{Type}} \\
\hline
\endfirsthead
\hline
\rowcolor{headerbg}
\textcolor{white}{\textbf{Round}} & \textcolor{white}{\textbf{Constraints}} & \textcolor{white}{\textbf{Type}} \\
\hline
\endhead
\endfoot

\rowcolor{lightblue}
\textbf{T1 ($N=3$)} &
\begin{enumerate}[left=0pt, nosep, label={[\theenumi]}]
  \item The response should include at least \textbf{4 sentences}.
  \item Highlight the word \textbf{performance} at least twice in the response.
  \item The answer must contain exactly \textbf{3 bullet points}.
\end{enumerate} &
Sentence count \newline Format/highlight \newline Bullet count \\
\hline

\rowcolor{lightyellow}
\textbf{T2 ($N=5$)} &
\begin{enumerate}[left=0pt, nosep, label={[\theenumi]}]
  \item Use specific rating terms such as \textbf{excellent}, \textbf{good}, \textbf{average}, \textbf{below average}, and \textbf{poor}.
  \item Write the response in a \textbf{formal business report style}.
  \item Limit the response to \textbf{50 words}.
  \item Include only \textbf{two sentences} in the response.
  \item Rate the employee's performance considering \textbf{teamwork} and \textbf{communication skills} in addition to the given tasks.
\end{enumerate} &
Rating vocabulary \newline Business register \newline Word limit \newline Sentence limit \newline Evaluation criteria expansion \\
\hline

\rowcolor{lightgreen}
\textbf{T3 ($N=5$)} &
\begin{enumerate}[left=0pt, nosep, label={[\theenumi]}]
  \item Use numerical values and specific rating terms such as \textbf{excellent}, \textbf{good}, \textbf{average}, \textbf{below average}, and \textbf{poor}.
  \item Limit the response to exactly \textbf{50 words}.
  \item Write in a \textbf{formal tone} suitable for a performance review document.
  \item Present the rating in \textbf{Markdown format}, including a table with columns for each activity and a final row for the overall performance rating.
  \item Use a role-based constraint: you are a \textbf{human resources manager} providing feedback.
\end{enumerate} &
Numerical rating \newline Exact word count \newline Formal review register \newline Markdown table format \newline Role-based evaluation \\
\hline
\end{longtable}

\textbf{Analysis:} T1 mainly tests basic format control through sentence count, repeated highlighting, and bullet count; T2 increases difficulty by adding rating terminology, business-report style, a tight word limit, two-sentence compression, and additional evaluation criteria such as teamwork and communication; T3 further raises complexity through an exact 50-word requirement, numerical scoring, Markdown table construction, activity-level comparison, and a human-resources-manager perspective.

\subsection{Case 8: Casino Royale CIA Plot Question Title}

\textbf{Task:} \texttt{Summarize a Casino Royale plot question about why the CIA let Le Chiffre remain free after Bond won the game.}

\begin{longtable}{|p{0.22\linewidth}|p{2.4in}|p{1.3in}|}
\hline
\rowcolor{headerbg}
\textcolor{white}{\textbf{Round}} & \textcolor{white}{\textbf{Constraints}} & \textcolor{white}{\textbf{Type}} \\
\hline

\rowcolor{lightblue}
\textbf{T1 ($N=5$)} &
\begin{enumerate}[left=0pt, nosep, label={[\theenumi]}]
  \item Summary must use at least one professional term related to \textbf{espionage or criminal activities}.
  \item Include no more than \textbf{two sentences} in the title.
  \item Tone should be \textbf{inquisitive and analytical}.
  \item Title should not exceed \textbf{20 words}.
  \item Title must be in the form of a \textbf{question}.
\end{enumerate} &
Professional terminology \newline Sentence limit \newline Tone constraint \newline Word limit \newline Interrogative form \\
\hline

\rowcolor{lightyellow}
\textbf{T2 ($N=5$)} &
\begin{enumerate}[left=0pt, nosep, label={[\theenumi]}]
  \item Use specific terms such as \textbf{`CIA'}, \textbf{`Bond'}, and \textbf{`Le Chiffre'} in the title.
  \item Limit the title to \textbf{15 words}.
  \item Write the title in the style of a \textbf{journalistic headline}.
  \item Include a reference to the \textbf{potential consequences} of Le Chiffre's actions.
  \item Ensure the title reflects \textbf{intrigue and curiosity}.
\end{enumerate} &
Named entities \newline Word limit \newline Headline style \newline Consequence reference \newline Curiosity tone \\
\hline

\rowcolor{lightgreen}
\textbf{T3 ($N=5$)} &
\begin{enumerate}[left=0pt, nosep, label={[\theenumi]}]
  \item Use the term \textbf{`CIA'} exactly \textbf{twice} in the title.
  \item Include the concept of \textbf{`trust betrayal'} in the title.
  \item Limit the title to \textbf{15 words}.
  \item Write the title in a \textbf{journalistic style}.
  \item Tailor the title for a \textbf{student audience}.
\end{enumerate} &
Exact lexical frequency \newline Concept inclusion \newline Word limit \newline Journalistic style \newline Audience constraint \\
\hline
\end{longtable}

\textbf{Analysis:} T1 asks for a short analytical question with espionage terminology; T2 adds named entities, journalistic headline style, and consequences; T3 increases precision by requiring `CIA' exactly twice, adding `trust betrayal', and tailoring the 15-word title to a student audience.

\subsection{Case 9: Computer Adjective Generation}

\textbf{Task:} \texttt{Generate a list of at least 5 adjectives to describe the given item. input: Computer}

\begin{longtable}{|p{0.22\linewidth}|p{2.4in}|p{1.3in}|}
\hline
\rowcolor{headerbg}
\textcolor{white}{\textbf{Round}} & \textcolor{white}{\textbf{Constraints}} & \textcolor{white}{\textbf{Type}} \\
\hline

\rowcolor{lightblue}
\textbf{T1 ($N=3$)} &
\begin{enumerate}[left=0pt, nosep, label={[\theenumi]}]
  \item The response should contain exactly \textbf{5 adjectives} to describe the given item.
  \item Your answer must not include any forbidden words such as \textbf{`slow'}, \textbf{`dumb'}, or \textbf{`primitive'}.
  \item The response should be in \textbf{English all lowercase letters}.
\end{enumerate} &
Exact count \newline Forbidden words \newline Lowercase English \\
\hline

\rowcolor{lightyellow}
\textbf{T2 ($N=3$)} &
\begin{enumerate}[left=0pt, nosep, label={[\theenumi]}]
  \item The response should contain exactly \textbf{5 adjectives} to describe the given item.
  \item Highlight each adjective with \textbf{asterisks} for emphasis and avoid using any \textbf{commas}.
  \item The response must be in \textbf{English}, and use the word \textbf{`fast'} at least once.
\end{enumerate} &
Exact count \newline Emphasis format \newline No commas \newline Mandatory word \\
\hline

\rowcolor{lightgreen}
\textbf{T3 ($N=5$)} &
\begin{enumerate}[left=0pt, nosep, label={[\theenumi]}]
  \item Must include the words \textbf{`high-speed'}, \textbf{`versatile'}, and \textbf{`innovative'} in your adjectives.
  \item Each adjective must relate to either \textbf{performance}, \textbf{design}, or \textbf{functionality} of a computer.
  \item Limit your response to exactly \textbf{15 words}.
  \item Use a \textbf{positive and enthusiastic tone} in your adjectives.
  \item Present the adjectives using \textbf{bullet points}.
\end{enumerate} &
Mandatory words \newline Semantic relevance \newline Word count \newline Positive tone \newline Bullet format \\
\hline
\end{longtable}

\textbf{Analysis:} T1 requires exactly five lowercase English adjectives while avoiding forbidden words; T2 adds emphasis formatting, comma avoidance, and mandatory use of the word \textbf{`fast'}; T3 further increases constraint complexity by requiring specific lexical items, semantic relevance to computer-related attributes, an exact word count, a positive tone, and bullet-point formatting.

\subsection{Case 10: Impact of Increasing Business Taxes}

\textbf{Task:} \texttt{Describe the impact of this policy. input: Increasing taxes for businesses.}

\begin{longtable}{|p{0.22\linewidth}|p{2.4in}|p{1.3in}|}
\hline
\rowcolor{headerbg}
\textcolor{white}{\textbf{Round}} & \textcolor{white}{\textbf{Constraints}} & \textcolor{white}{\textbf{Type}} \\
\hline
\endfirsthead
\hline
\rowcolor{headerbg}
\textcolor{white}{\textbf{Round}} & \textcolor{white}{\textbf{Constraints}} & \textcolor{white}{\textbf{Type}} \\
\hline
\endhead
\endfoot

\rowcolor{lightblue}
\textbf{T1 ($N=5$)} &
\begin{enumerate}[left=0pt, nosep, label={[\theenumi]}]
  \item Must include the phrase \textbf{economic burden} in the response.
  \item Highlight the effects on \textbf{small businesses} and \textbf{large businesses} separately.
  \item Write the description in a \textbf{formal tone} suitable for a business journal article.
  \item Limit the response to \textbf{100 words}.
  \item Divide the response into \textbf{two paragraphs}.
\end{enumerate} &
Mandatory phrase \newline Stakeholder separation \newline Formal register \newline Word limit \newline Paragraph structure \\
\hline

\rowcolor{lightyellow}
\textbf{T2 ($N=5$)} &
\begin{enumerate}[left=0pt, nosep, label={[\theenumi]}]
  \item Mandate the use of the terms \textbf{revenue}, \textbf{profit}, and \textbf{investment}.
  \item Limit the response to \textbf{three sentences}.
  \item Write from the perspective of an \textbf{economist} highlighting both positive and negative impacts.
  \item Use a \textbf{formal and professional tone}.
  \item Present the information in \textbf{Markdown format}, with headers for main points.
\end{enumerate} &
Economic terminology \newline Sentence limit \newline Expert perspective \newline Balanced analysis \newline Markdown headers \\
\hline

\rowcolor{lightgreen}
\textbf{T3 ($N=5$)} &
\begin{enumerate}[left=0pt, nosep, label={[\theenumi]}]
  \item Mandate the use of the terms \textbf{economic burden}, \textbf{revenue generation}, and \textbf{market competitiveness}.
  \item Limit the response to \textbf{four sentences}.
  \item Write in a \textbf{formal and professional tone} suitable for a policy analyst report.
  \item Output the description in \textbf{Markdown format}, including headings and bullet points for key impacts.
  \item Consider the perspectives of \textbf{small}, \textbf{medium}, and \textbf{large businesses} separately.
\end{enumerate} &
Policy terminology \newline Sentence limit \newline Analyst register \newline Markdown headings and bullets \newline Multi-stakeholder comparison \\
\hline
\end{longtable}

\textbf{Analysis:} T1 requires a formal two-paragraph description that distinguishes small and large business impacts while including the phrase \textbf{economic burden}; T2 shifts to a more analytical economist perspective, adds required economic terms, demands balanced positive and negative impacts, and introduces Markdown headers; T3 further increases difficulty by requiring policy-analysis terminology, Markdown headings and bullets, and separate treatment of small, medium, and large businesses, making the stakeholder analysis more granular and demanding.

\section{Limitations}
\label{sec:limitations}

A limitation of our work is that real-world user instructions can be substantially more complex than our training data. Users may write instructions thousands of tokens long, interweave multiple constraint types, reference external documents, and include implicit requirements that are only loosely implied rather than explicitly stated. The AgentIF benchmark~\citep{qi2025agentif} illustrates this challenge, with instructions averaging 1,723 words and nearly 12 constraints each. Nevertheless, SEIF remains effective on AgentIF, suggesting its potential for handling complex instruction-following scenarios. As model capabilities continue to improve, future work can explore evolving more realistic and complex instructions to further enhance models' ability to follow complex user requirements.

\section{Crowdsourcing and research with human subjects}
\label{sec:Crowdsourcing}
We recruit three computer science students as annotators. The annotators are paid above the local minimum wage and consent to the use of the annotated data for the research purpose covered in this paper. When annotations conflict, we adopt majority voting to determine the final label.

\section{Declaration of LLM usage}
\label{sec:llm}

We acknowledge the use of \href{https://github.com/cursor/cursor}{Cursor} as an AI-assisted writing tool. It was used solely for language polishing of an early version of the manuscript. All core ideas presented in this paper were independently developed by the authors.

\section{Impact}
\label{sec:impact}
SEIF explores a self-evolving paradigm for improving large language models. In this paradigm, models can continuously construct training challenges and learn from their own evolving experience. This direction may contribute to more scalable and autonomous post-training methods, reducing the long-term dependence on intensive human supervision or increasingly stronger teacher models. More broadly, this work suggests a path toward adaptive AI systems that can improve on open-ended tasks through iterative interaction between data generation, quality control, reward modeling, and policy optimization. Such a paradigm may benefit future research on model alignment, lifelong learning, curriculum learning, and autonomous capability development.

\end{document}